\documentclass[10pt,twocolumn,letterpaper]{article}
\makeatletter
\providecommand{\@LN}[2]{}
\makeatother
\usepackage[accsupp]{axessibility}  

\usepackage[pagenumbers]{cvpr} 

\usepackage{color}

\newcommand{\nc}{n_{\mathrm{c}}}

\newcommand \R {\mathbb{R}}
\newcommand{\mean}{\operatorname{mean}} 
\newcommand{\std}{\operatorname{std}} 
\newcommand{\Gram}{\operatorname{Gram}} 

\definecolor{cvprblue}{rgb}{0.21,0.49,0.74}
\usepackage[pagebackref,breaklinks,colorlinks,allcolors=cvprblue]{hyperref}

\def\paperID{7157} 
\def\confName{CVPR}
\def\confYear{2025}

\title{SGSST: Scaling Gaussian Splatting Style Transfer}

\author{
\parbox{\textwidth}{\centering Bruno Galerne$^{1,2}$, Jianling Wang$^{1}$,
Lara Raad$^{3}$,
and Jean-Michel Morel$^{4}$
        }
        \\
\normalsize\parbox{\textwidth}{\smallskip\centering $^1$Université d'Orléans, Université de Tours, CNRS, IDP, UMR 7013, Orléans, France\\
$^2$Institut Universitaire de France (IUF) \\
$^3$Instituto de Ingeniería El\'ectrica, Facultad de Ingeniería, Universidad de la Rep\'ublica
\quad
$^4$City University of Hong Kong
}
}

\usepackage{array}
\usepackage{multirow}
\usepackage{tabularx}

\newcommand{\PreserveBackslash}[1]{\let\temp=\\#1\let\\=\temp} 
\newcolumntype{C}[1]{>{\PreserveBackslash\centering}p{#1}}
\newcolumntype{R}[1]{>{\PreserveBackslash\raggedleft}p{#1}}
\newcolumntype{L}[1]{>{\PreserveBackslash\raggedright}p{#1}}

\begin{document}
\twocolumn[{%
\renewcommand\twocolumn[1][]{#1}%
\maketitle
\begin{center}
    \centering
    \captionsetup{type=figure}
    \includegraphics[width=\textwidth]{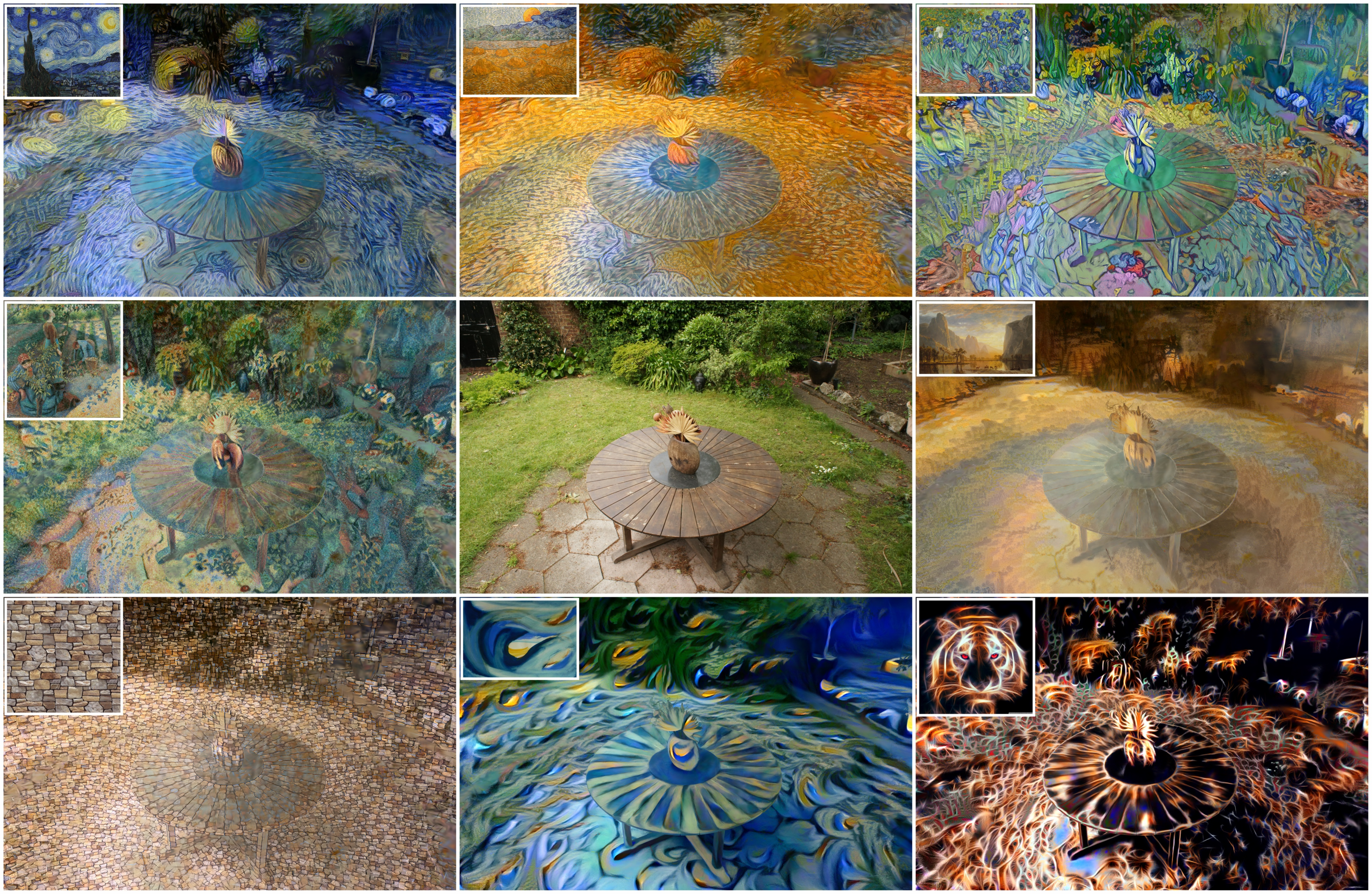}
    \captionof{figure}{\textbf{Various ultra-high definition style transfers of a Gaussian splatting 3D scene.} 
    SGSST transfers a very large set of global style statistics of an image to a 3DGS scene by minimizing a tailored multiscale SOS loss, yielding 3D style transfer of superior quality and at unprecedented high resolution (images have size 5187$\times$3361).}
    \label{fig:teaser}
\end{center}%
}]

\begin{abstract}
   Applying style transfer to a full 3D environment is a challenging task that has seen many developments since the advent of neural rendering.
   3D Gaussian splatting (3DGS) has recently pushed further many limits of neural rendering in terms of training speed and reconstruction quality.
   This work introduces SGSST: Scaling Gaussian Splatting Style Transfer, an optimization-based method to apply style transfer to pretrained 3DGS scenes.
   We demonstrate that a new multiscale loss based on global neural statistics, that we name SOS for Simultaneously Optimized Scales, enables style transfer to ultra-high resolution 3D scenes. Not only SGSST pioneers 3D scene style transfer at such high image resolutions, it also produces superior visual quality as assessed by thorough qualitative, quantitative and perceptual comparisons.
\end{abstract}

\section{Introduction}
\label{sec:intro}

Dealing with ultra-high resolution (UHR) rendering is capital for AR/VR applications. 
Indeed, when navigating in a 3D environment the user only sees a partial field of view of the environment.
This adds challenging issues for applying style transfer to 3D environments, that is, transferring the visual characteristics of an image such as a painting to a 3D scene.
Indeed, while the general scene should convey the global color palette of the style painting, when getting closer to objects in a stylized environment the user should expect to see fine painting patterns such as brushstrokes.
However, in the current state of the art, the user quickly encounters resolution-limited content in the form of blurry interpolated features.

Ever since the seminal work of Mildenhall \emph{et al.}~\cite{Mildenhall_etal_NeRF_ECCV2022}, 
neural radiance fields (NeRF) have seen many improvements. Several 3D scene representations have been proposed for improving the quality and resolution of the reconstructed scenes as well as easing the training, e.g.~ \cite{Fridovich-Keil_Yu_etal_plenoxels_radiance_fields_without_neural_networks_CVPR2022, Barron_etal_Mip-NeRF360_unbounded_antialiased_neural_radiance_fields_CVPR2022, Muller_etal_instant_neural_graphics_primitives_with_a_multiresolution_hash_encoding_SIGGRAPH2022, Chen_etal_TensorRF_tensorial_radiance_fields_ECCV2022}.
The recent 3D Gaussian splatting (3DGS)~\cite{Kerbl_etal_3D_Gaussian_splatting_for_real-time_radiance_field_rendering_SIGGRAPH2023} 
has introduced an efficient  high-resolution (HR) scene representation which has stirred much interest~\cite{Wu_etal_recent_advances_in_3D_Gaussian_splatting_CVM2024}.
Both frameworks have stirred attempts to 3D style transfer algorithms~\cite{Chiang_etal_stylizing_3D_scene_via_implicit_representation_and_hyperNetwork_WACV2022, Huang_etal_StylizedNeRF_Consistent_3D_Scene_Stylization_As_Stylized_NeRF_via_2D-3D_CVPR2022, Zhang_etal_arf_artistic_radiance_fields_ECCV2022, Liu_etal_StyleRF_zero_shot_3D_style_transfer_of_neural_radiance_fields_CVPR2023,Saroha_etal_Gaussian_splatting_in_style_ArXiv2024,Liu_etal_StyleGaussian_instant_3D_style_transfer_with_Gaussian_splatting_ArXiv2024, Zhang_etal_stylisedGS_controllable_stylization_for_3D_Gaussiansplatting_ArXiv2024, Kovacs_etal_G_style_stylized_gaussian_splatting_CGF2024}, 
but these methods have so far produced medium resolution outputs. They do not faithfully transport high resolution multiscale textures such as those present in paintings.
Motivated by a recent neural style transfer (NST) \cite{Gatys_et_al_image_style_transfer_cnn_cvpr2016} contribution tailored for UHR images \cite{Galerne_etal_scaling_painting_style_transfer_EGSR2024}, i.e. with resolution larger than 4k images, we show  that 3DGS can be leveraged for UHR style transfer.

The contributions of this work are the following:

\begin{itemize}
    \item We introduce SOS, a Simultaneously Optimized Scales loss expressed in a single parameterless and explainable formula.
    \item By solely optimizing the SOS loss, we reach UHR for 3DGS style transfer and we scale Gaussian Splatting Style Transfer by a four times resolution gain.
    \item Superior quality transfer: By transferring a very large set of global style statistics, we obtain superior style transfer quality even at HR resolution, as confirmed by a comparative  perceptual study. 
\end{itemize}

In short, our approach is the first method that allows UHR style transfer directly to 3DGS.
It produces high visual quality results and relies on optimizing a single multiscale loss.
The simplicity of our approach ensures its reproducibility.
Being optimization-based, SGSST's main limitation is a fairly large training time that is two to eight times longer than the initial 3DGS training depending on the image resolution.
Yet, considering the high quality of the results and their reproducibility, 
this contribution is valuable for AR/VR applications that require high visual quality for their user experience.

\section{Related work}
\label{sec:related_works}

\paragraph{Neural style transfer}
In the seminal work of Gatys \emph{et al.}~\cite{Gatys_et_al_image_style_transfer_cnn_cvpr2016} NST is formulated as an optimization problem minimizing the distances between Gram matrices of VGG~\cite{Simonyan_Zisserman_very_deep_cnn_vgg_ICLR2015} features. 
Even though other VGG statistics have been considered, almost all subsequent style transfer and texture synthesis methods rely on VGG~\cite{Sendik_deep_correlations_texture_synthesis_SIGGRAPH2017, gonthier2022high, Vacher_etal_texture_interpolation_probing_visual_perception_NEURIPS2020, Lu_Zhu_Wu_Deepframe_AAAI2016, DeBortoli_et_al_maximum_entropy_methods_texture_synthesis_SIMODS2021, Risser_etal_stable_and_controllable_neural_texture_synthesis_and_style_transfer_Arxiv2017, Heitz_slices_Wassestein_loss_neural_texture_synthesis_CVPR2021}. 
To accelerate style transfer, several methods ~\cite{ulyanov2016texturenets,Ulyanov_etal_improved_texture_networks_CVPR2017,johnson2016Perceptual} have attempted to train feed-forward networks approximately minimizing the Gram loss~\cite{Gatys_et_al_image_style_transfer_cnn_cvpr2016}. 
However, these approximate methods  require learning a new model for each style type. 
This latter limitation has been addressed by fast Universal Style Transfer (UST) approaches~\cite{chen2016fast,Huang_arbitrary_style_transfer_real_time_ICCV2017,li2017universal,sheng2018avatar,park2019arbitrary,li2019learning,chiu2020iterative} that use VGG feature decoders. 

For HR images, coarse-to-fine multiscale strategies \cite{Gatys_etal_Controlling_perceptual_factors_in_neural_style_transfer_CVPR2017, Snelgrove_hr_multiscale_neural_texture_synthesis_SIGGRAPHAsia2017, gonthier2022high} have  proved effective, but still face limitations due the high GPU memory usage of VGG statistics.
Fast HR alternatives~\cite{an2020real,Wang_2020_CVPR,Wang_etal_MicroAST_AAAI2023,Chen_Wang_Xie_Lu_Luo_towards_ultra_resolution_neural_style_transfer_thumbnail_instance_normalization_AAAI2022,Texler_etal_arbitrary_style_transfer_using_neurally_guided_patch_synthesis_CG2020} do exist but
generally suffer from artifacts and struggle to capture the full style complexity.
Recently, SPST~\cite{Galerne_etal_scaling_painting_style_transfer_EGSR2024} proposed an implementation of the Gatys \emph{et al.} method adapted to UHR 
 (larger that 4k) images. The visual quality of SPST's results is superior, at the cost of a long optimization time.

\paragraph{Style transfer for neural radiance fields}
NeRFs~\cite{Mildenhall_etal_NeRF_ECCV2022} have completely redefined the field of 3D scene modeling and novel view synthesis.
Editing the visual aspect of NeRFs via style transfer has quickly been addressed
~\cite{Chiang_etal_stylizing_3D_scene_via_implicit_representation_and_hyperNetwork_WACV2022, Huang_etal_StylizedNeRF_Consistent_3D_Scene_Stylization_As_Stylized_NeRF_via_2D-3D_CVPR2022, Zhang_etal_arf_artistic_radiance_fields_ECCV2022, Liu_etal_StyleRF_zero_shot_3D_style_transfer_of_neural_radiance_fields_CVPR2023}, usually by fine tuning a pretrained NeRF representation using a style transfer loss, or training an additional fast style transfer module.
ARF~\cite{Zhang_etal_arf_artistic_radiance_fields_ECCV2022} is a notable exception: It uses Nearest Neighbor Feature Matching (NNFM) for fine tuning a plenoxel radiance field~\cite{Fridovich-Keil_Yu_etal_plenoxels_radiance_fields_without_neural_networks_CVPR2022},
produces high-quality results at moderate resolution, and is the base model for other  methods~\cite{Zhang_etal_coARF_controllable_3D_artistic_style_transfer_for_radiance_fields_3DV2024, Jung_etal_geometry_transfer_for_stylizing_radiance_fields_CVPR2024}.
While these works paved the way for radiance field style transfer, they are all limited in input and output image resolution.

\paragraph{Style transfer for Gaussian splatting}
A few recent contributions show that 3DGS is a promising framework for 3D scene style transfer.
Saroha \emph{et al} \cite{Saroha_etal_Gaussian_splatting_in_style_ArXiv2024} propose a solution for universal style transfer of a given 3DGS scene. 
The method processes the colors of Gaussians with a tiny MLP trained using fast AdaIN \cite{Huang_arbitrary_style_transfer_real_time_ICCV2017} and relying on a multi-resolution hash grid representation \cite{Muller_etal_instant_neural_graphics_primitives_with_a_multiresolution_hash_encoding_SIGGRAPH2022}.
StyleGaussian~\cite{Liu_etal_StyleGaussian_instant_3D_style_transfer_with_Gaussian_splatting_ArXiv2024} is a concurrent approach that relies on transferring encoding of VGG features to each Gaussian and applying AdaIN to these features. The new Gaussian features are then decoded into an RGB color by processing the K-nearest neighbors of the Gaussians.
After training, both methods allow for instantaneous stylization with any style image, but the visual quality of the results is quite low, since high-resolution details such as brushstroke patterns are not transferred.

StylizedGS~\cite{Zhang_etal_stylisedGS_controllable_stylization_for_3D_Gaussiansplatting_ArXiv2024} is an optimization-based method that extends ARF~\cite{Zhang_etal_arf_artistic_radiance_fields_ECCV2022} to 3DGS. It uses a training loss made of six terms combined with a color transfer preprocessing, the loss being designed to minimize changes in the 3DGS geometry while letting the style evolve via VGG NN matching.
$\mathcal{G}$-Style~\cite{Kovacs_etal_G_style_stylized_gaussian_splatting_CGF2024}
 also is an optimization based approach that further uses a CLIP-based loss~\cite{Radford_CLIP_ICML2021}.
While these approaches produce slightly  better results than ARF~\cite{Zhang_etal_arf_artistic_radiance_fields_ECCV2022}, both methods are limited in content and style resolution due to the use of nearest-neighbor matching of VGG features.

None of the current state of the art deals with HR style transfer.
The stylization is either fast but very approximate due to AdaIn \cite{Saroha_etal_Gaussian_splatting_in_style_ArXiv2024, Liu_etal_StyleGaussian_instant_3D_style_transfer_with_Gaussian_splatting_ArXiv2024} or unable to use HR style images due to NNFM~\cite{Zhang_etal_arf_artistic_radiance_fields_ECCV2022, Zhang_etal_stylisedGS_controllable_stylization_for_3D_Gaussiansplatting_ArXiv2024, Kovacs_etal_G_style_stylized_gaussian_splatting_CGF2024}.
To the best of our knowledge, SGSST is the first procedure allowing high-quality transfer for 3DGS that is trained and rendered at UHR.

\section{Preliminaries}

\subsection{Gaussian splatting representation}

Starting from a multiview training set of images $\{u_i\}_{i=1}^{N_{\mathrm{views}}}$ of a static scene accompanied with corresponding calibrated cameras $\{\mathcal{C}_i\}_{i=1}^{N_{\mathrm{views}}}$ computed by structure from motion~\cite{Schonberger_and_Frahm_structure_from_motion_revisited_CVPR2016},
the 3DGS algorithm~\cite{Kerbl_etal_3D_Gaussian_splatting_for_real-time_radiance_field_rendering_SIGGRAPH2023} trains a set of colored 3D Gaussians 
$\{\mathcal{G}_j\}_{j=1}^{N_{\mathrm{Gaussians}}}$ so that they represent the 3D scene from any camera position.
Each Gaussian $\mathcal{G}_j$ is represented by a finite set of features: a center position $\mu_j$, a covariance matrix $\Sigma_j$ (encoded by a scaling diagonal matrix and a rotation matrix), an opacity $\alpha_j$ and a view-dependent color function $c_j$.
These parameters are used in a volumetric renderer that determines the color by summing the contribution of each Gaussian that intersects a ray with direction $(\theta, \varphi)$ via $\alpha$-blending (see e.g. \cite{Mildenhall_etal_NeRF_ECCV2022}).
The contribution of a Gaussian is its color $c_j(\theta, \varphi)$ multiplied by an opacity $\sigma_j$ defined as the maximal opacity $\alpha_j$ times the unnormalized Gaussian density at the ray position \cite{Kerbl_etal_3D_Gaussian_splatting_for_real-time_radiance_field_rendering_SIGGRAPH2023}.
Thus, the resulting color $C$ for the ray is 
\begin{equation}
C = \sum_{j=1}^N c_j(\theta, \varphi) \sigma_j \prod_{k=1}^{j-1} (1-\sigma_k)
\end{equation}
where the sum is over all Gaussians intersecting the ray.
The color function $c_j(\theta, \varphi)$ associated with a Gaussian depends on the spherical direction $(\theta, \varphi)$ through an order 3 spherical harmonics polynomial function given by
\begin{equation}
c_j(\theta, \varphi) = c_{j,0} + \sum_{\ell=1}^3 \sum_{m=1}^{2\ell+1} c_{j,\ell, m} Y_{\ell, m}(\theta, \varphi)
\end{equation}
where the vectors $c_{j,0}$ and $c_{j,\ell, m}$ are in $\R^3$ and the $Y_{\ell, m}$ form a basis of the spherical harmonics polynomials of degree $\ell$.
In short, $c_{j,0}\in\R^3$ is the main color and the additional coefficients $c_{j,\ell, m}$ encode  smooth variations of this color when the viewing angle changes.

Like for NeRF, the key ingredient of the 3DGS parametrization is the differentiable rendering function $\mathcal{R}(\mathcal{C}; \Theta)$ that renders a view of the scene for any camera $\mathcal{C}$ given the scene parameters 
$\Theta = \{ (\mu_j, \Sigma_j, \alpha_j, c_{j,0}, (c_{j,\ell, m})_{\ell,m})\}_{j=1}^{N_{\mathrm{Gaussians}}}$. 
This differentiable rendering function allows to train the Gaussian parameters to minimize the reconstruction error
\begin{equation}
\min_{\Theta}\frac{1}{N_{\mathrm{views}}} \sum_{i=1}^{N_{\mathrm{views}}} E_{\mathrm{reconstruction}} (\mathcal{R}(\mathcal{C}_i; \Theta); u_i)
\label{eq:min_reconstruction_3dgs}
\end{equation}
where $E_{\mathrm{reconstruction}}$ is a 2D image comparison error (such as a combination of mean square error and SSIM \cite{Kerbl_etal_3D_Gaussian_splatting_for_real-time_radiance_field_rendering_SIGGRAPH2023}).
The minimization is conducted using Adam~\cite{Kingma_Ba_Adam_ICLR2015} for 30k iterations by randomly picking one view at each iteration.

\subsection{Style transfer loss for UHR images}

\begin{figure*}[ht!]
\includegraphics[width=\linewidth]{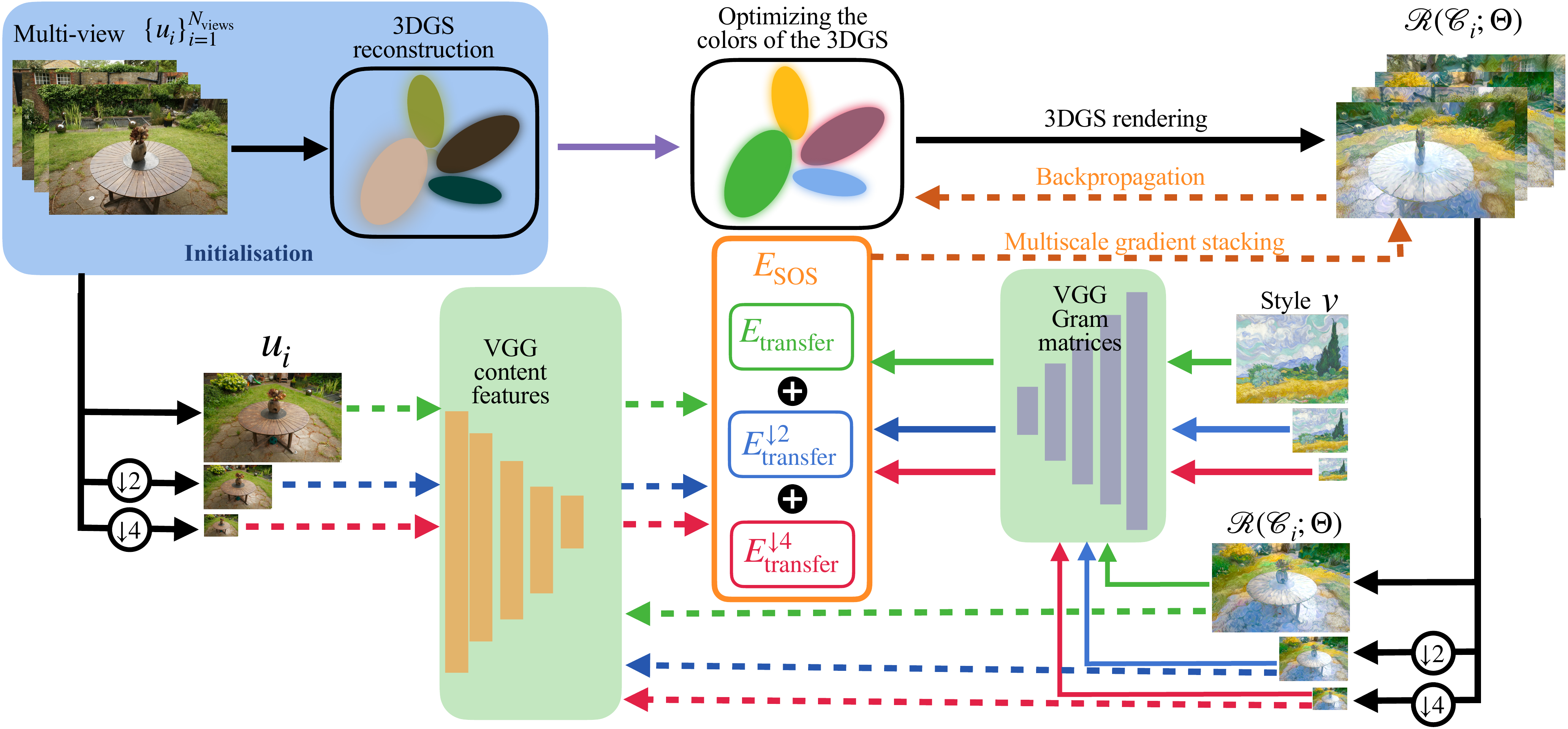}
\caption{\textbf{Overview of SGSST.} Starting from a pretrained realistic 3DGS scene \cite{Kerbl_etal_3D_Gaussian_splatting_for_real-time_radiance_field_rendering_SIGGRAPH2023}, we optimize the colors of each Gaussian using the new multiscale SOS loss (involving $n_{\mathrm{s}}=3$ scales in the illustration).
Computing the gradient w.r.t. the loss is feasible for UHR images thanks to the SPST partition-based implementation~\cite{Galerne_etal_scaling_painting_style_transfer_EGSR2024}.
Multiscale gradient stacking is used at the node of the rendered image to perform only one backpropagation per iteration through the 3DGS rendering pipeline.}
\label{fig:pipeline}
\end{figure*}

Our approach relies on optimizing VGG19~\cite{Simonyan_Zisserman_very_deep_cnn_vgg_ICLR2015} feature statistics as initially proposed by Gatys \emph{et al.}~\cite{Gatys_et_al_image_style_transfer_cnn_cvpr2016}.
Content consistency is loosely monitored by preservation of the feature layer
$L_\mathrm{c} = \mathtt{ReLU\_4\_2}$ while style transfer is imposed by matching spatial statistics of five VGG19 layers, namely the set 
$\mathcal{L}_\mathrm{s} = \{\mathtt{ReLU\_k\_1},~k\in\{1,2,3,4,5\}\}$.
The statistics of interest of the feature response $V^L(w)$ of an image $w$ at some layer $L\in\mathcal{L}_\mathrm{s}$ 
having $\nc^L$ feature channels are
its Gram matrix $\Gram(V^L(w)) \in \R^{\nc^L\times \nc^L}$,
its spatial mean vector $\mean(V^L(w)) \in \R^{\nc^L}$, and its standard deviation vector $\std(V^L(w))\in \R^{\nc^L}$.
Given a content image $u$ and a style image $v$, we consider the loss function
\begin{equation}
E_{\mathrm{transfer}}(x;u,v) 
= 
E_{\mathrm{content}}(x;u) 
+ E_{\mathrm{style}}(x;v)
\label{eq:gatys_loss_texture_transfer}
\end{equation}
where 
$E_{\mathrm{content}}(x;u) = \lambda_\mathrm{c} \left\| V^{L_\mathrm{c}}(x) -  V^{L_\mathrm{c}}(u)\right\|^2$, with $\lambda_\mathrm{c}>0$
and the style loss is defined by  
\begin{equation}
E_{\mathrm{style}}(x;v) = \sum_{L \in \mathcal{L}_\mathrm{s}} E_{\mathrm{style}}^L(x;v)
\label{eq:style_loss}.
\end{equation}
where $E_{\mathrm{style}}^L(x;v)$ is a linear combination of the mean square error between the VGG19 statistics of $V^L(x)$ and the one of the style features $V^L(v)$~\cite{Galerne_etal_scaling_painting_style_transfer_EGSR2024}.
While only the Gram matrices were originally used~\cite{Gatys_et_al_image_style_transfer_cnn_cvpr2016}, it has been shown that adding control for the mean and standard deviation corrects some style transfer color artefacts~\cite{Galerne_etal_scaling_painting_style_transfer_EGSR2024} previously identified in the literature~~\cite{Sendik_deep_correlations_texture_synthesis_SIGGRAPH2017,  Risser_etal_stable_and_controllable_neural_texture_synthesis_and_style_transfer_Arxiv2017, Heitz_slices_Wassestein_loss_neural_texture_synthesis_CVPR2021}.

To obtain high-quality style transfer for HR images one needs to optimize the style transfer loss using several scales and a coarse-to-fine approach~\cite{Gatys_etal_Controlling_perceptual_factors_in_neural_style_transfer_CVPR2017}. Indeed, if one applies style transfer on the highest resolution only, the changes within the content image are limited to local texture and the results does not convey a painting aspect.
Due to the use of VGG19 features, computing the loss $E_{\mathrm{transfer}}(x;u,v)$ and its gradient with respect to (w.r.t.) $x$ via backpropagation is memory prohibiting for UHR images. 
However, using a grid partition and local loss backpropagation based on precomputed global statistics, SPST~\cite{Galerne_etal_scaling_painting_style_transfer_EGSR2024} allows for the exact evaluation of this gradient.

\section{Scaling Gaussian splatting style transfer}

A complete overview of our SGSST algorithm is given in Figure~\ref{fig:pipeline}.
Starting from a realistic 3DGS representation~\cite{Kerbl_etal_3D_Gaussian_splatting_for_real-time_radiance_field_rendering_SIGGRAPH2023}, we optimize the colors of each Gaussian using a multiscale style transfer loss.

\subsection{Stylizing the 3DGS representation}

By changing the reconstruction loss of Equation \eqref{eq:min_reconstruction_3dgs} with a style transfer loss for the input style image $v$, one can hope to stylize a realistic 3DGS.
However, given the complexity of the 3DGS representation and the many interacting parameters, it is not such an easy task to alter the 3DGS aspects without loosing the content geometry.
Our solution is to only optimize for the constant color components $\Theta_{\mathrm{color}} = \{c_{j,0}\}_{j=1}^{N_{\mathrm{Gaussians}}}$ of the Gaussians and simply freeze all the other parameters $\Theta_{\mathrm{init.}}$ given by the initial realistic 3DGS training.

We experimentally found that this robust approach ensures a rich style transfer and fully preserves the 3D geometry.
Indeed, fixing all the Gaussian parameters except for the main color component $c_{j,0}$ preserves the scene geometry, as the location and size of the Gaussians are being kept (see Section~\ref{subsec:ablation_studies} for ablation experiments).

\subsection{Multiscale Style Transfer Loss}
\label{subsec:multiscale_style_transfer_loss}

\begin{figure*}[ht!]
\includegraphics[width=\textwidth]{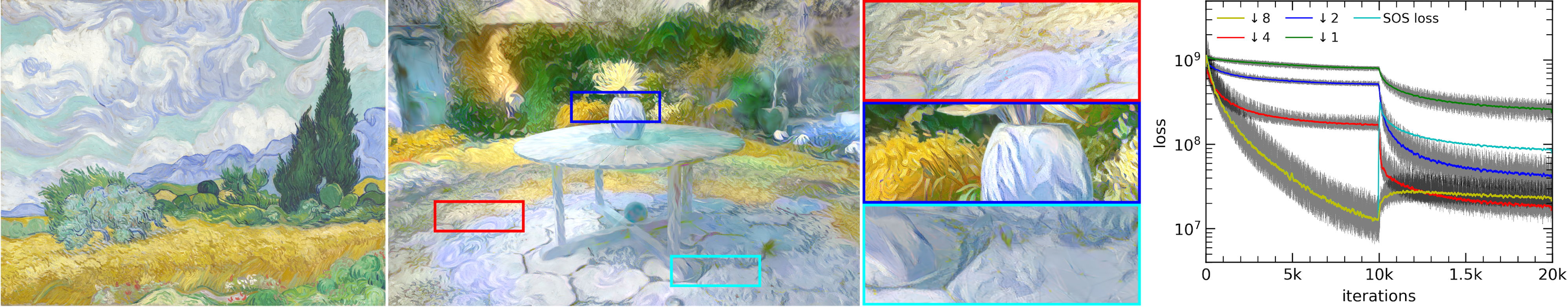}
\caption{\textbf{UHR 3DGS style transfer.} SGSST allows for the multiscale style transfer of 3DGS scenes at UHR. From left to right: Style image, one UHR stylized view, three magnified details, and evolution of the SOS loss and each style transfer loss that contributes to it.
We first optimize the transfer loss for the coarsest scale (yellow curve) for 10k iterations and then optimize for another 10k iterations the SOS loss (light blue curve), namely the mean of the four transfer losses. Images sizes are 5187$\times$3361 for content and 4230$\times$3361 for style.}
\label{fig:uhr_result}
\end{figure*}

We introduce the Simultaneously Optimized Scales (SOS) loss
defined as
\begin{equation}
E_{\mathrm{SOS}}(x;u,v) = \frac{1}{n_{\mathrm{s}}} \sum_{s=0}^{n_{\mathrm{s}}-1}
E_{\mathrm{transfer}}(x^{\downarrow 2^s};u^{\downarrow 2^s},v^{\downarrow 2^s}) 
\label{eq:multiscale_loss}
\end{equation}
where $n_{\mathrm{s}}\geq 1$ is the number of considered scales and $u^{\downarrow 2^s}$ denotes  an image $u$ downscaled by a  $2^s$ factor.
This SOS loss \eqref{eq:multiscale_loss} enables style transfer simultaneously at all  scales. 
A somewhat similar approach was proposed for 2D texture synthesis~\cite{Snelgrove_hr_multiscale_neural_texture_synthesis_SIGGRAPHAsia2017} but, as already mentioned, multiscale 2D style transfer is generally conducted using a coarse-to-fine strategy~\cite{Gatys_etal_Controlling_perceptual_factors_in_neural_style_transfer_CVPR2017}.
However, for 3DGS we observed that the initial configuration had only little influence on the final result, making the coarse-to-fine strategy ineffective for multiscale style transfer
(See Section~\ref{subsec:ablation_studies}).

In the end, as illustrated by Figure~\ref{fig:pipeline}, the stylization of UHR 3DGS is conducted by solving for
\begin{equation}
\min_{\Theta_{\mathrm{color}}}
\frac{1}{N_{\mathrm{views}}} \sum_{i=1}^{N_{\mathrm{views}}} E_{\mathrm{SOS}}( \mathcal{R}(\mathcal{C}_i ; \Theta); u_i, v),
\label{eq:min_multiscale_loss_3dgs}
\end{equation}
where $\Theta$ stands for the 3DGS parameters obtained by replacing the color components of $\Theta_{\mathrm{init.}}$ by the values of the optimization variable $\Theta_{\mathrm{color}}$.

\subsection{Implementation details}
\label{subsec:implementation_details}

\paragraph{Multiscale gradient stacking}
The 2D SPST method \cite{Galerne_etal_scaling_painting_style_transfer_EGSR2024} provides the gradient of each term $E_{\mathrm{transfer}}(x^{\downarrow 2^s};u^{\downarrow 2^s},v^{\downarrow 2^s})$ w.r.t. to the input $x^{\downarrow 2^s} = \mathcal{R}(\mathcal{C}_i ; \Theta)^{\downarrow 2^s}$.
We first backpropagate each of these gradients through the downscaling operator and stack them at the level of the rendered image $\mathcal{R}(\mathcal{C}_i ; \Theta)$.
When each of the $n_{\mathrm{s}}$ scales has been treated, the stacked gradient is equal to the gradient of the full loss $E_{\mathrm{SOS}}( \mathcal{R}(\mathcal{C}_i ; \Theta); u_i, v)$ w.r.t. to the rendered image $\mathcal{R}(\mathcal{C}_i ; \Theta)$. 
This gradient is then backpropagated through the Gaussian rendering pipeline to obtain the gradient w.r.t. the Gaussian colors $\Theta_{\mathrm{color}}$ (see both orange arrows in Figure~\ref{fig:pipeline}).
Using this strategy we only backpropagate one time per iteration through the 3DGS rendering pipeline instead of $n_{\mathrm{s}}$ times.

\paragraph{Color transfer via style transfer at the coarsest scale} 
To speed up the color transfer, we first optimize for 10k Adam iterations with the loss restricted to the coarsest scale 
$E_{\mathrm{transfer}}(\mathcal{R}(\mathcal{C}_i ; \Theta)^{\downarrow 2^s};u_i^{\downarrow 2^s},v^{\downarrow 2^s})$ with $s = n_{\mathrm{s}}-1$, then optimize the SOS loss for another 10k Adam iterations.

\paragraph{Number of scales} 
$n_{\mathrm{s}}$ is set automatically to use all available scales, the coarsest resolution having sides larger than 256 for VGG19 statistics to be reliable. 

\paragraph{Reproducibility}
All our experiments have been conducted using the same SOS loss and training procedure, making our approach parameterless and fully reproducible. 
Our public PyTorch implementation is based on the public source codes\footnote{\url{https://github.com/graphdeco-inria/gaussian-splatting}; \url{https://github.com/bgalerne/scaling_painting_style_transfer}} for 3DGS~\cite{Kerbl_etal_3D_Gaussian_splatting_for_real-time_radiance_field_rendering_SIGGRAPH2023} training and SPST~\cite{Galerne_etal_scaling_painting_style_transfer_EGSR2024}.
Our code and videos of our results are available online\footnote{Code: \url{https://github.com/JianlingWANG2021/SGSST}; 

Videos: \url{https://www.idpoisson.fr/galerne/sgsst/}}.

\section{Experiments}

\begin{figure*}
\includegraphics[width=\textwidth]{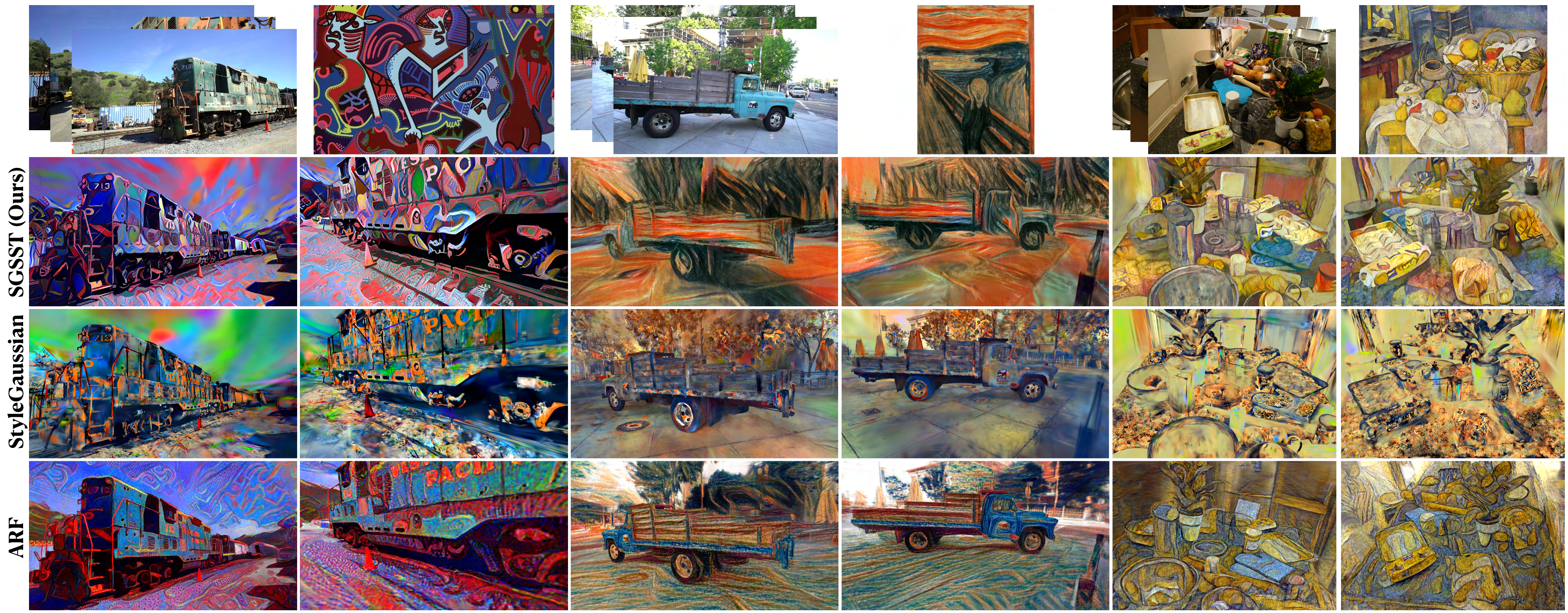}
\caption{\textbf{Comparison of SGSST (ours, top) with StyleGaussian~\cite{Liu_etal_StyleGaussian_instant_3D_style_transfer_with_Gaussian_splatting_ArXiv2024} (middle) and ARF~\cite{Zhang_etal_arf_artistic_radiance_fields_ECCV2022} (bottom).}
 From left to right the content resolutions are 
980$\times$545 (train), 
979$\times$546 (truck), 
and 
3115$\times$2076 (counter).
For the first two  examples, the various outputs keep the resolution of the content, but for the HR counter scene, the output sizes are 3115$\times$2076 for SGSST, 1600$\times$1066 for StyleGaussian and 779$\times$519 for ARF (see supp. mat. Figure 28 for ARF results without downscaling).
Thanks to its multiscale global VGG statistics, SGSST is the most faithful method regarding style consistency.%
} 
\label{fig:comparison}
\end{figure*}

\subsection{Ultra-high resolution results}
\label{sec:uhr-results}

Our multiscale stylization algorithm is able to transfer style details at UHR for both the content image resolution and the style image resolution.
This results in unprecedentedly rich style transfer, as illustrated by Figure~\ref{fig:uhr_result}.
As can be observed, minimizing the SOS loss indeed allows to decrease the style transfer loss for all scales (Figure~\ref{fig:uhr_result} right).
The approach is especially relevant when transferring the style of an UHR painting presenting style features at several scales, ranging from a specific  color palette to a main curve style and to fine brushstrokes or canvas texture (see close-up views of Figure~\ref{fig:uhr_result} and the first two lines of Figure~\ref{fig:teaser}).
To obtain such results, combining style transfer at the largest possible number of scales is critical (see ablation in supp. mat. Figure 9). 
In addition, the approach is also efficient to transfer  the style of a medium resolution style image to an UHR scene, as shown in the last row of Figure~\ref{fig:teaser}.

To the best of our knowledge, our  approach is the first to work at UHR resolution for neural style transfer of neurally rendered 3D scenes.
One of the main advantages of the Gram-based loss is that it does not depend on the style resolution and,  when both the content and style images grow in $O(N)$, 
its complexity scales in $O(N)$ while NNFM used in ARF~\cite{Zhang_etal_arf_artistic_radiance_fields_ECCV2022} scales in $O(N^2)$.

Yet, applying style transfer at such resolutions remains computationally heavy: for the garden image of size 5187$\times$3361 (Figures~\ref{fig:teaser} and~\ref{fig:uhr_result}) the style transfer takes 25.5 hours (VS 3 hours to train the initial 3DGS model), that is an 8.5 overhead factor. For an image of moderate size (Figure~\ref{fig:comparison} left and middle) the style transfer and the initial 3DGS training take respectively 22 and 10 minutes, that is, the overhead factor is only 2.2. 
These time values were obtained using a single A100 GPU with 80 GB of memory and could be accelerated by adapting the SOS loss implementation to a multi-GPU setting. 
Note that high computation times are inherent to UHR style transfer: Running the 2D SPST method for the 185 garden training images takes 27 hours (8.9 min. per image).

\subsection{Comparison}

We perform a thorough comparative study using 40 3D style transfer experiments using 9 different scenes and various style images (see supp. mat.).
We compare our results with the NeRF-based ARF~\cite{Zhang_etal_arf_artistic_radiance_fields_ECCV2022} and 
the 3DGS-based StyleGaussian~\cite{Liu_etal_StyleGaussian_instant_3D_style_transfer_with_Gaussian_splatting_ArXiv2024} algorithms, described in Section~\ref{sec:related_works}, using their public implementations\footnote{\url{https://github.com/Kai-46/ARF-svox2}; \url{https://github.com/Kunhao-Liu/StyleGaussian}}.

Comparing style transfer methods is challenging because each algorithm treats the style image differently.
Following high quality 2D style transfer~\cite{Gatys_etal_Controlling_perceptual_factors_in_neural_style_transfer_CVPR2017, Galerne_etal_scaling_painting_style_transfer_EGSR2024}, our loss uses up to four scales and each scale uses five VGG layers.
The UHR style images were downscaled so that the style has the same size as the content images (no upscale was applied if the style image is smaller than the content image).
In contrast, ARF uses a single VGG layer from a single scale and the style image is downscaled to be twice smaller than the content input, resulting in smaller local texture.
StyleGaussian reduces the style images so that it has the size 256$\times$256, a scale that is hardly sufficient to represent HR style images such as paintings.

\paragraph{Qualitative comparison}

Figure~\ref{fig:comparison} shows three different comparative experiments.
As one can observe, the results of StyleGaussian~\cite{Liu_etal_StyleGaussian_instant_3D_style_transfer_with_Gaussian_splatting_ArXiv2024} are generally not satisfying since the method fails to transfer local texture or to preserve the style image's color palette.
ARF results better reproduce brushstroke textures, but the transfer is limited, as the method only involves a single VGG layer at a single scale. Also, NNFM does not ensure the preservation of a global color palette.
In comparison, SGSST    preserves the style at all considered scales. This results in color palette preservation, as well as a verifiable transfer of features of any size, from large brushstrokes (Figure~\ref{fig:comparison} middle) to local grain transfer (Figure~\ref{fig:comparison} right).
In addition, SGSST is the only method that performs style transfer at the original resolution of the HR example of Figure~\ref{fig:comparison} right.

\paragraph{Quantitative comparison}

Even though there is no consensus for the quantitative evaluation of NST~\cite{Ioannou_Maddock_evaluation_in_neural_style_transfer_a_review_CGF2024}, following previous works, we report two different metrics for style transfer quality and texture consistency across views.
Style transfer quality can be measured by 
the Gram loss~\cite{Gatys_et_al_image_style_transfer_cnn_cvpr2016}.
A second metric proper to NeRF style transfer~\cite{Liu_etal_StyleRF_zero_shot_3D_style_transfer_of_neural_radiance_fields_CVPR2023, Liu_etal_StyleGaussian_instant_3D_style_transfer_with_Gaussian_splatting_ArXiv2024, Saroha_etal_Gaussian_splatting_in_style_ArXiv2024} is to 
check the short-term and long-term consistency of the radiance fields in terms of LPIPS~\cite{Zhang_etal_LPIPS_CVPR2018} and RMSE between wrapped views. 
Since SGSST is the only method working with UHR images, when necessary we forcefully downgraded all the results to the resolution of ARF for comparison.
The average of these two metrics over our 40 experiments is reported in Table~\ref{tab:metrics}.
The Gram loss is the best for SGSST and, surprisingly, StyleGaussian achieves a lower Gram loss than ARF, which is not consistent with the qualitative evaluation, probably explained by ARF using a single VGG layer compared to five for the two other methods.
In terms of consistency metrics, StyleGaussian reports to be the most stable approach followed by SGSST but note that this metric favors the lack of texture.

\begin{table}[t]    
    \centering
    \small
    \begin{tabular}{@{}L{0.18\linewidth}C{0.14\linewidth}C{0.105\linewidth}C{0.105\linewidth}C{0.105\linewidth}C{0.105\linewidth}@{}}
        \toprule
        Method & 
        \multicolumn{1}{c}{\parbox{0.14\linewidth}{\centering Transfer quality}} & 
        \multicolumn{2}{c}{\parbox{0.21\linewidth}{\centering Short-range consistency}} & 
        \multicolumn{2}{c}{\parbox{0.21\linewidth}{\centering Long-range consistency}} \tabularnewline
        \midrule
        & Gram$\downarrow$ & LPIPS$\downarrow$ & RMSE$\downarrow$ & LPIPS$\downarrow$ & RMSE$\downarrow$ \tabularnewline
        \midrule
        SGSST & $\mathbf{2.59\mathrm{e}8}$ & $\mathbf{0.030}$  & \underline{$0.032$} & \underline{$0.055$} & \underline{$0.063$} \tabularnewline
        StyleGaussian & \underline{$4.61\mathrm{e}8$} & \underline{$0.033$} & $\mathbf{0.029}$ & $\mathbf{0.050}$ & $\mathbf{0.056}$ \tabularnewline
        ARF & $5.77\mathrm{e}8$ & $0.040$ & $0.037$ & $0.072$ & $0.066$ \tabularnewline
        \bottomrule
    \end{tabular}
    \caption{\textbf{Quantitative comparison.} Style transfer quality and texture consistency metrics averaged over 40 experiments for SGSST, StyleGaussian, and ARF.
        Best results in bold, second best underlined.}
    \label{tab:metrics}
\end{table}

\paragraph{Perceptual study}

To further support our results, we conducted a perceptual study comparing the 40 3D style transfer experiments.
For each example, four images were displayed (at a resolution of 1280$\times$720): the style image and, in a blind random order, the results of the three methods (SGSST, ARF and StyleGaussian) shown at a common viewpoint (also randomly selected).
Each participant was shown ten random instances and was asked to select the result that was the most faithful to the style image.
The study was conducted on the web with volunteer participants.

A total of 68 participants took part in the test, resulting in 680 votes summarized in Table~\ref{tab:perceptual-study}. 
This perceptual study shows that the fast style transfer performed by StyleGaussian is consistently judged inferior in quality over ARF and SGSST. 
It also confirms that SGSST is far superior to ARF in terms of visual quality since 66\% of the participants ranked our method first for its  painting style transfer quality, even when presented with results downscaled in resolution.

\begin{table}[t]
    \centering
    \small
    \begin{tabularx}{0.47\textwidth}{@{\extracolsep{\fill}}cccc@{}}
    \toprule
    & SGSST & ARF & StyleGaussian \tabularnewline
    \midrule    
    Voting results (\%) & \textbf{66.3} & \underline{31.6} & 2.1 \tabularnewline
    \bottomrule
    \end{tabularx}
    \caption{
    \textbf{Perceptual study.} Summary of the {680} votes for the most style consistent result.
    }
    \label{tab:perceptual-study}
\end{table}

\subsection{Ablation study}
\label{subsec:ablation_studies}

\paragraph{Influence of optimization parameters}

\begin{figure}
\includegraphics[width=\linewidth]{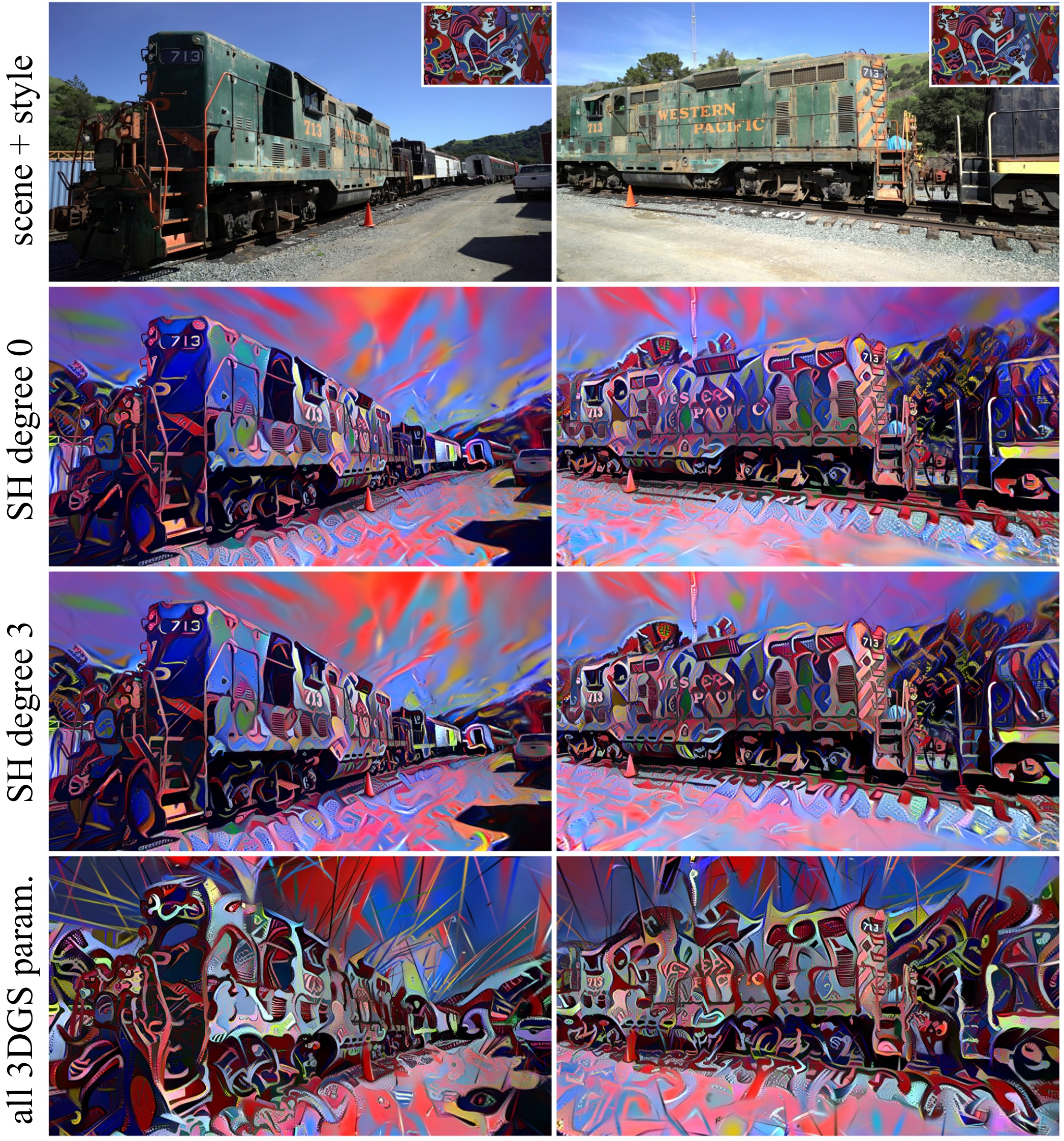}
\caption{\textbf{Influence of optimization parameters:} Allowing more 3DGS parameters to be optimized when minimizing the SOS loss does not improve the stylization quality and can dramatically impact the geometry. From left to right: Style image, content, SGSST default (optimization of colors), results when optimizing all spherical harmonics, results when optimizing all parameters.}
\label{fig:influence_optimization_parameters}
\end{figure}

Our SGSST algorithm stylizes a realistic 3DGS scene by tuning the constant color components of the 3DGS Gaussians using a single 2D style transfer loss function at multiple scales (Equation~\eqref{eq:multiscale_loss}).
In contrast, Zhang \emph{et al.}~\cite{Zhang_etal_stylisedGS_controllable_stylization_for_3D_Gaussiansplatting_ArXiv2024} 
optimize all 3DGS parameters. 
This, however, necessitates  a complex loss made of six different terms to avoid artefacts: 
a loss term enforces consistency with the original geometry via depth preservation while a preprocessing of Gaussians floaters is necessary after color transfer.

Figure~\ref{fig:influence_optimization_parameters} shows that optimizing more parameters of the 3DGS for SGSST brings no benefit.
Optimizing all the spherical harmonic coefficients does not improve the result, and letting all the parameters free like in~\cite{Zhang_etal_stylisedGS_controllable_stylization_for_3D_Gaussiansplatting_ArXiv2024, Kovacs_etal_G_style_stylized_gaussian_splatting_CGF2024} leads to a strong degradation of the geometry.
Note that two instantaneous 3DGS style transfer methods~\cite{Saroha_etal_Gaussian_splatting_in_style_ArXiv2024, Liu_etal_StyleGaussian_instant_3D_style_transfer_with_Gaussian_splatting_ArXiv2024} are based on modifying the color features via some neural networks, but their results are not comparable to optimization-based approaches in terms of visual quality.

\begin{figure*}
\includegraphics[width=\linewidth]{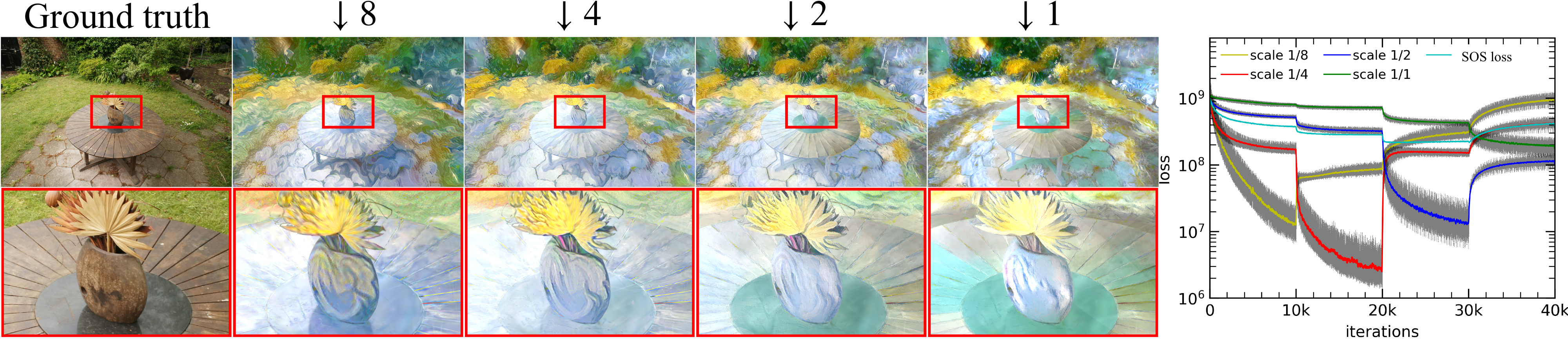}
\caption{\textbf{Failure of coarse-to-fine strategy.} 
Results of coarse-to-fine style transfer for the example of Figure~\ref{fig:uhr_result}. Each scale is initialized with the output of the previous scale. As shown by the evolution of the losses (right), when training at a new scale, the loss of the previous scales increases quickly. This explains why the large scale painting features disappear progressively and are absent after training the target UHR (see close-up details).}
\label{fig:coarse_to_fine} 
\end{figure*}

\paragraph{Failure of coarse-to-fine strategy} 
Our approach minimizes a style transfer loss simultaneously at all scales.
This is different from the coarse-to-fine style transfer strategy that has proven successful for HR style transfer~\cite{Gatys_etal_Controlling_perceptual_factors_in_neural_style_transfer_CVPR2017}.
Figure~\ref{fig:coarse_to_fine} illustrates that this coarse-to-fine strategy fails in the context of 3DGS. Indeed, the gain obtained by optimizing at a given scale is quickly lost when optimizing the next one leading to the disappearance of large scale features. 
This can be explained by the fact that the 3DGS representation is a more constrained representation than pixel grids due to its sparsity.

\begin{figure}

\includegraphics[width=\linewidth]{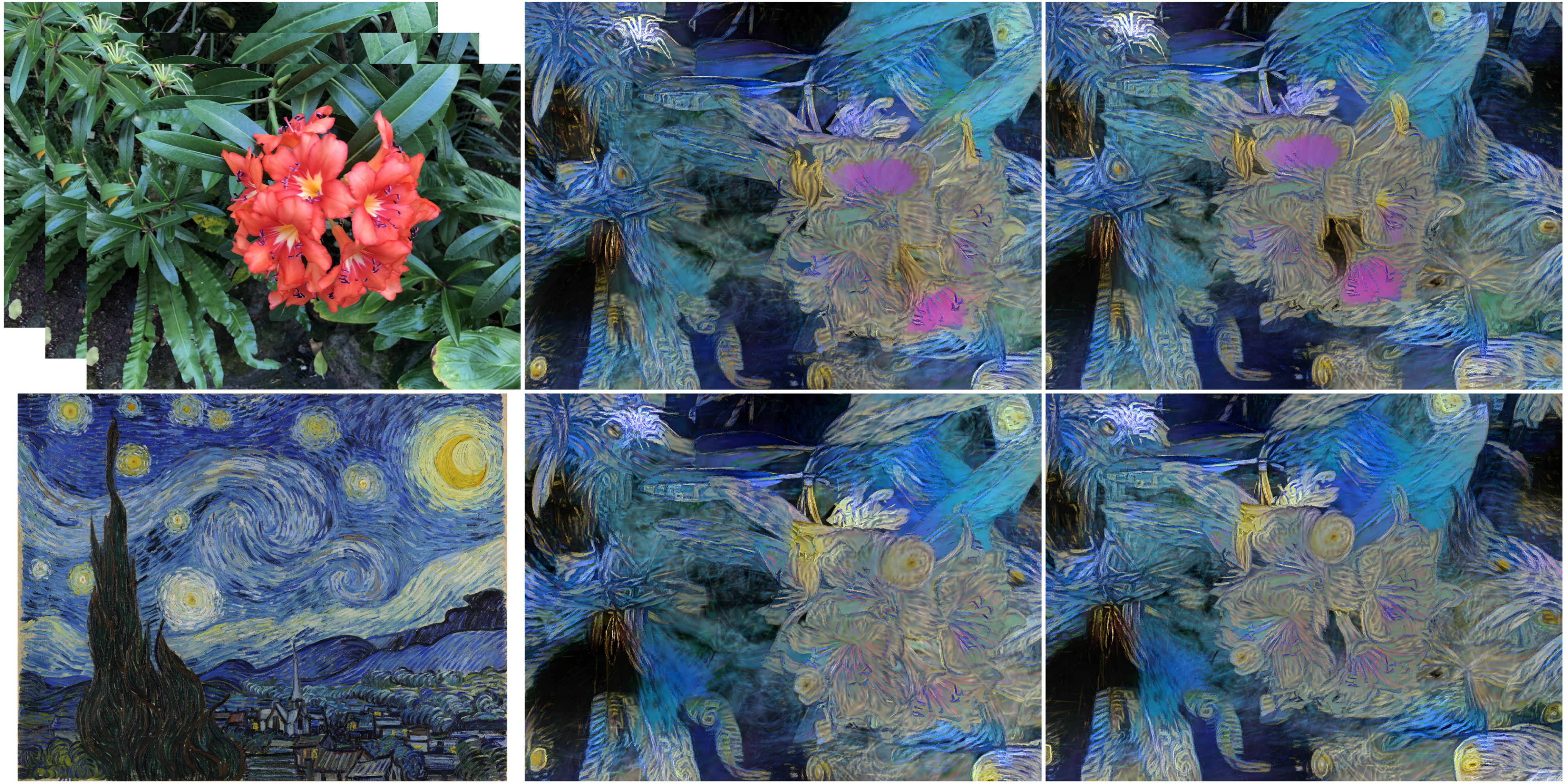}
\caption{\textbf{Color transfer via style transfer at the coarsest scale.} The first 10k iterations at the coarsest scale allow for quick color changes in the 3DGS scene.
Removing this step may lead to color artifacts in the result (top) compared to our default two-step optimization (bottom).}
\label{fig:initial_coarse_resolution}
\end{figure}

\paragraph{Color transfer via style transfer at the coarsest scale}

As described in Section~\ref{subsec:implementation_details}, we first optimize the style transfer loss at the lowest resolution for 10k iterations and then optimize the SOS loss for another 10k iterations. 
Figure~\ref{fig:initial_coarse_resolution} illustrates that these first iterations are  necessary for a faithful style color palette reproduction.

\section{Discussion and limitations}

\begin{figure}
\includegraphics[width=\linewidth]{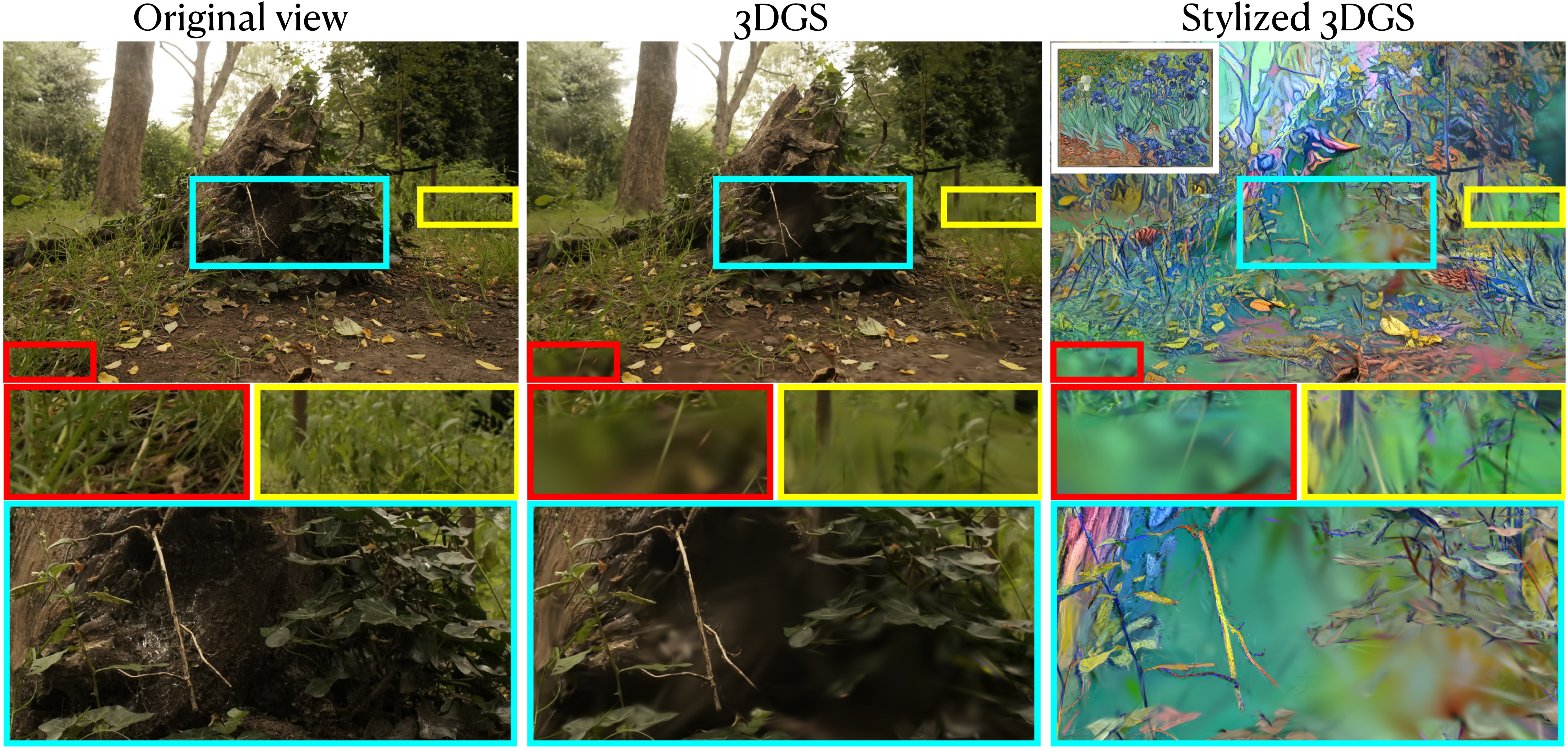}
\caption{\textbf{Limited texture representation due to low Gaussian density of the initial 3DGS.} From left to right: Original scene, 3DGS reconstruction, stylization of the 3DGS scene.}
\label{fig:resolution_limitation}
\end{figure}

\paragraph{Texture representation}
The texture representation within a 3DGS scene depends on the density of Gaussians and may be limited in low density areas. Isolated Gaussians can sometimes be spotted as illustrated by Figure~\ref{fig:resolution_limitation}.

\paragraph{Large computation time}
Depending on the resolution, SGSST requires from several minutes to several hours of computation. 
On the other hand, fast 3DGS stylization approaches~\cite{Saroha_etal_Gaussian_splatting_in_style_ArXiv2024, Liu_etal_StyleGaussian_instant_3D_style_transfer_with_Gaussian_splatting_ArXiv2024} do not reach a satisfactory visual quality.

\paragraph{Content-style mismatch}
As said earlier, the Gram loss has the advantage of being independent of the style's image resolution.
Also, it enables a faithful transfer of global statistics of the style image, such as its color palette. 
This important feature is, nevertheless, counterproductive when the style and content images strongly mismatch, leading to color bleeding or texturing of flat areas. 
Other controls can be added to mitigate these artefacts~\cite{Gatys_etal_Controlling_perceptual_factors_in_neural_style_transfer_CVPR2017} and it was shown that these controls are effective for 3DGS scenes~\cite{Zhang_etal_stylisedGS_controllable_stylization_for_3D_Gaussiansplatting_ArXiv2024}.

\section{Conclusion}

In this work we presented SGSST, a method that, for the first time, enables UHR 3DGS style transfer. To that aim, among other innovations, we introduced  the simultaneously optimized scales (SOS) loss. Our qualitative, quantitative and perceptual studies show that SGSST obtains superior style transfer quality than state of the art, even after reducing our results' resolution for a fair comparison with methods that do not reach UHR. Such high quality UHR results necessitate a large computation time that, nevertheless, remains comparable with that of UHR 3DGS training.

This work opens the way to several research directions. 
A first challenge is to  produce equally high quality style transfer with a faster algorithm based on UST.
A second more exploratory direction is to investigate geometry style transfer for 3DGS by designing adapted regularization to avoid the caveats depicted by Figure~\ref{fig:influence_optimization_parameters}.

 \clearpage

{\noindent\textbf{Acknowledgements:} B. Galerne and L. Raad acknowledge the support of the project MISTIC (ANR-19-CE40-005).
}

{
    \small
    \bibliographystyle{ieeenat_fullname}
    \bibliography{biblio_gaussian_splatting}
}

\cleardoublepage
\maketitlesupplementary

\appendix

\paragraph{Supplementary material description}
Our supplementary material consists of the following elements:
\begin{itemize}
    \item The present document with additional details and figures.
    \item The project website:
    \url{https://www.idpoisson.fr/galerne/sgsst/}
    with rendered videos, including one video showing the stylized scenes of the main paper's teaser and videos for the 40 comparison experiments (see Figures~\ref{fig:comparison_garden} to \ref{fig:comparison_truck}).
    \item The source code used for all experiments availble at \url{https://github.com/JianlingWANG2021/SGSST} based on the public source codes\footnote{\url{https://github.com/graphdeco-inria/gaussian-splatting}; \url{https://github.com/bgalerne/scaling_painting_style_transfer}} for 3DGS~\cite{Kerbl_etal_3D_Gaussian_splatting_for_real-time_radiance_field_rendering_SIGGRAPH2023} training and SPST~\cite{Galerne_etal_scaling_painting_style_transfer_EGSR2024}.
\end{itemize}

\smallskip

Note that due to space constraints all the images of this document have been compressed.

\section{Ablation on the number of scales}

\begin{figure*}
\includegraphics[width=\linewidth]{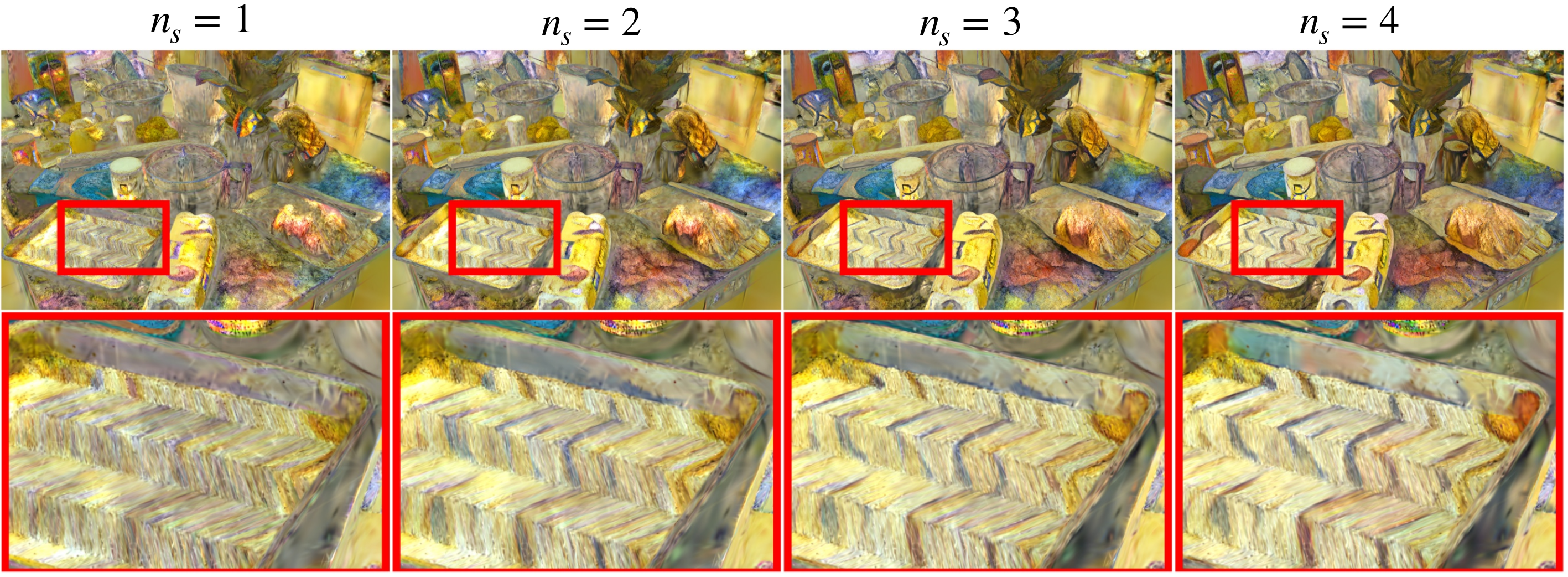}
\caption{\textbf{Ablation of the number of scales} of the SOS loss. 
Style transfer results using different number of scales (starting from the same initialization obtained by 10k iterations using coarsest scale for all).
High-quality style transfer is only achieved when using all scales ($n_{\mathrm{s}}=4$).}
\label{fig:ablation-nscales}
\end{figure*}

As explained in the main paper, the number of scales $n_{\mathrm{s}}$ is set automatically to use all available scales, the coarsest resolution having sides larger than 256 for VGG19 statistics to be reliable.
Figure~\ref{fig:ablation-nscales} presents an ablation on the number of scales $n_s$ showing the results for different values of $n_s$ and the corresponding close-ups of these results (after an initial color transfer for first 10k iterations using coarsest scale $n_s=4$ for all examples). 
One can observe that when using only the large resolution images ($n_s=1$) the pattern of the style transfer are limited to HR details.
High-quality style transfer is only achieved when using all scales.

\section{UHR style transfers of the teaser figure}

\begin{figure*}
\begin{center}
\includegraphics[height=100mm]{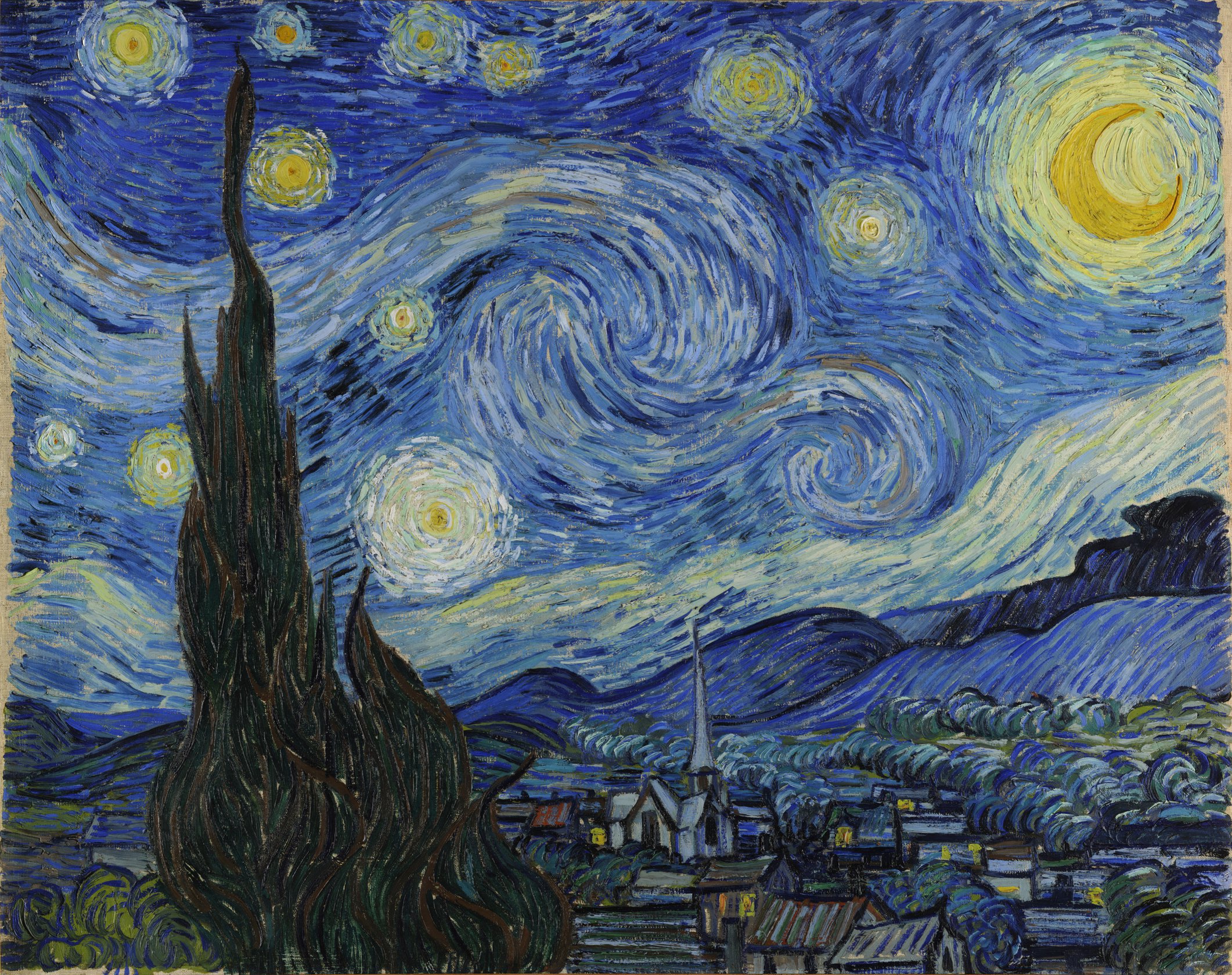}\\
\includegraphics[height=100mm]{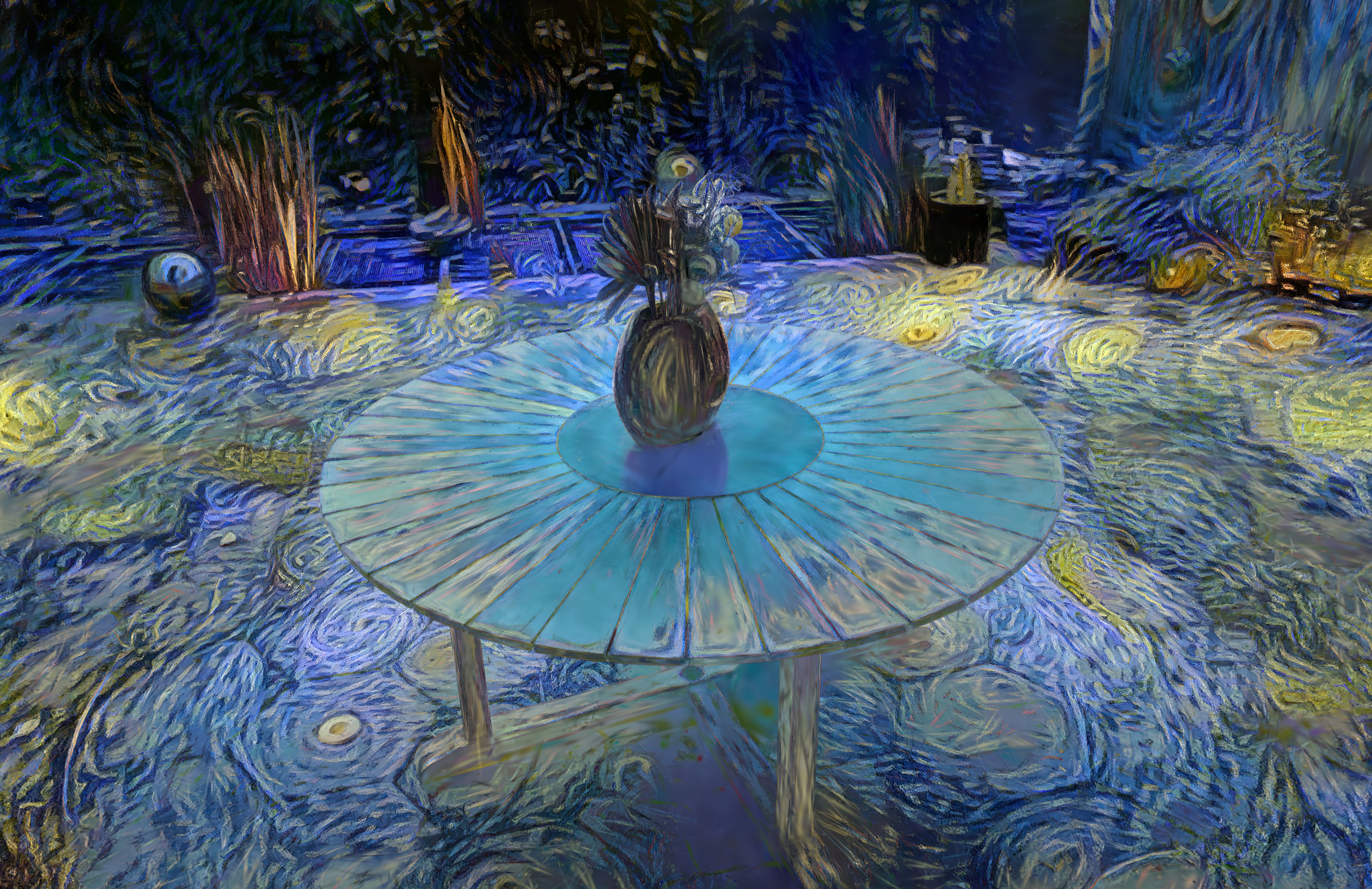}
\end{center}
\caption{Full view display of the example 1/8 of the teaser figure with the style image (size 4244$\times$3361) displayed at the same scale as the rendered image (size 5187$\times$3361). 
Images have been downscaled by a factor 2 and compressed using jpeg.}
\label{fig:supp_mat_teaser_full_1}
\end{figure*}

\begin{figure*}
\begin{center}
\includegraphics[height=100mm]{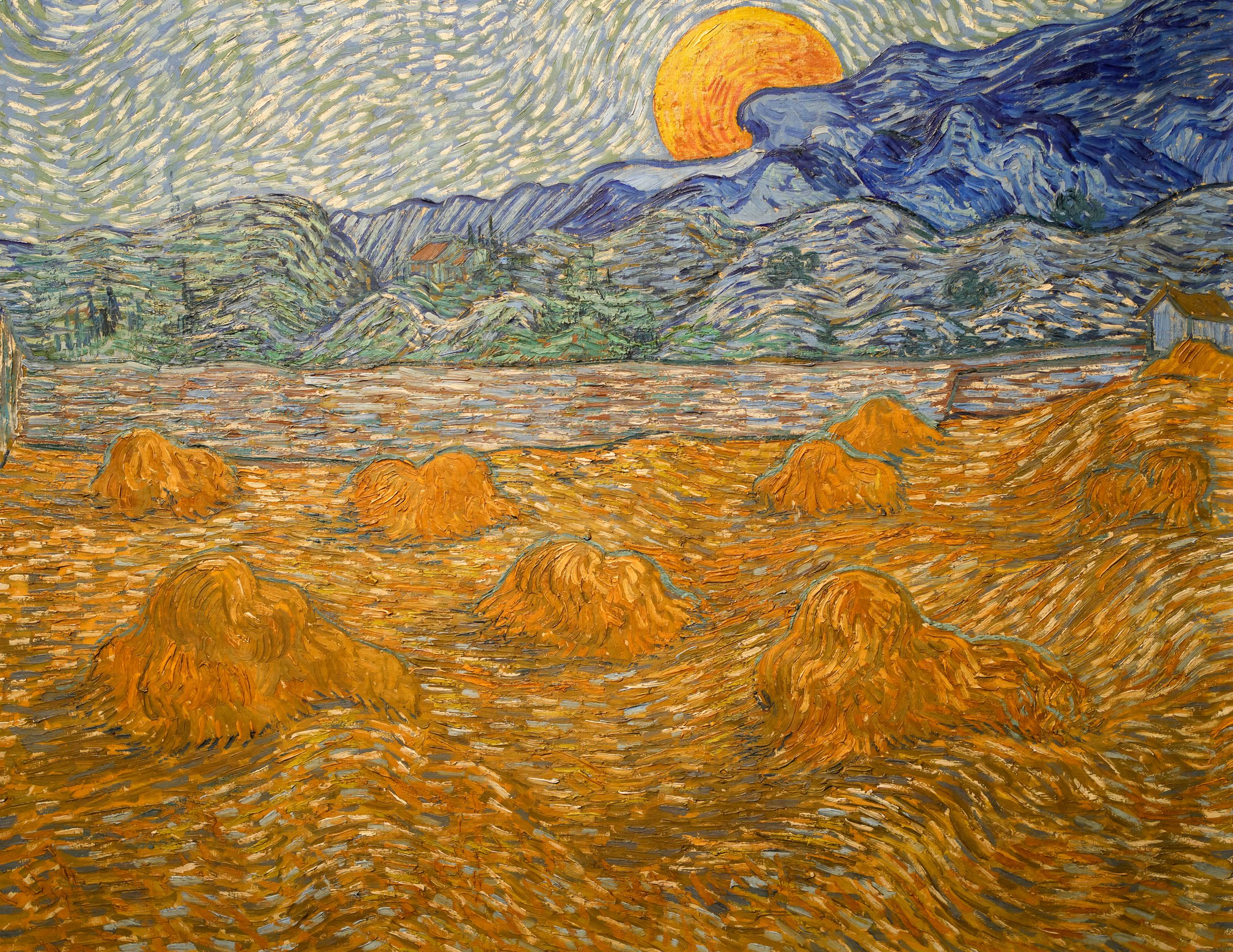}\\
\includegraphics[height=100mm]{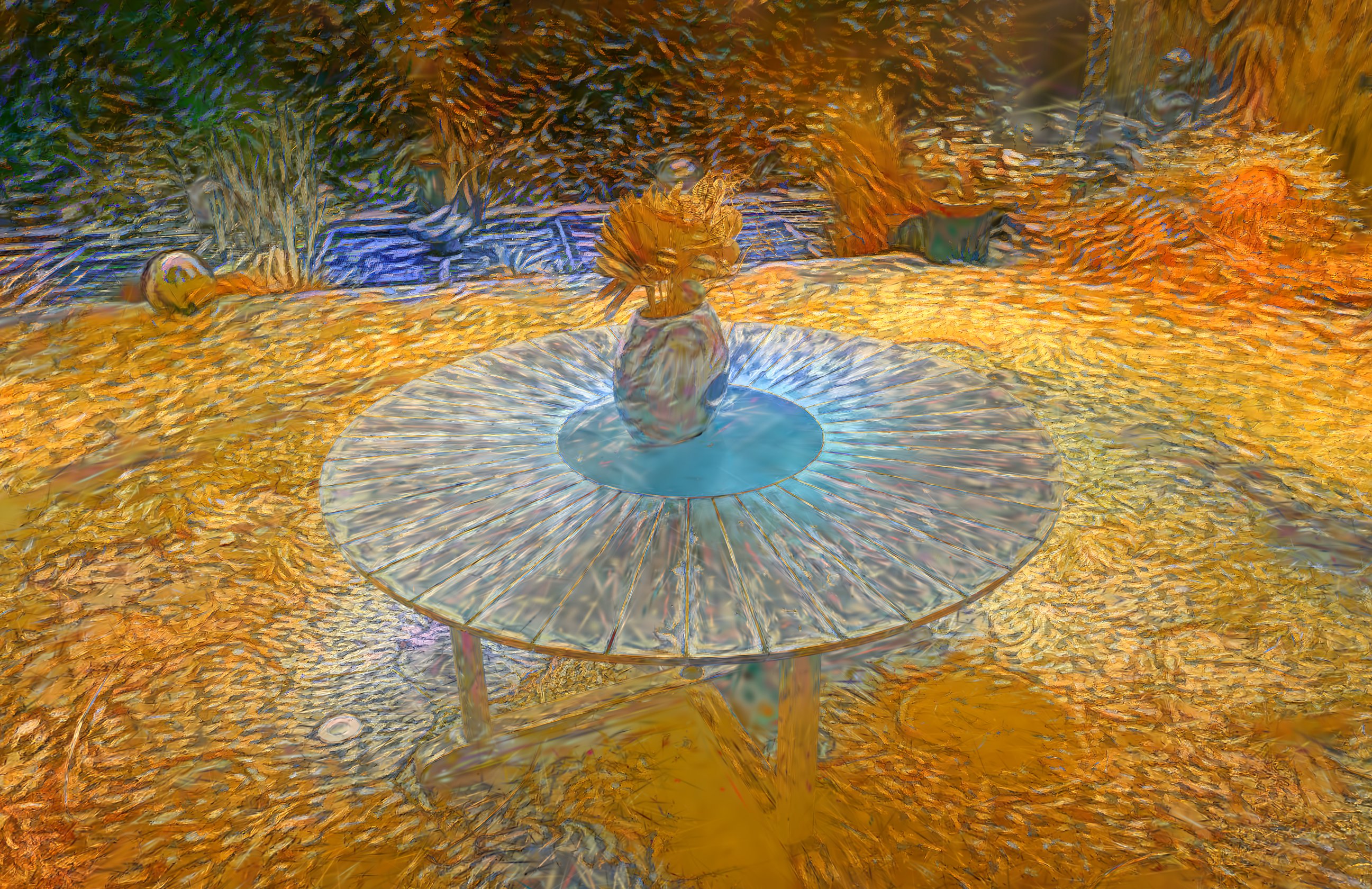}
\end{center}
\caption{Full view display of the example 2/8 of the teaser figure with the style image (size 4351$\times$3361) displayed at the same scale as the rendered image (size 5187$\times$3361). 
Images have been downscaled by a factor 2 and compressed using jpeg.}
\label{fig:supp_mat_teaser_full_2}
\end{figure*}

\begin{figure*}
\begin{center}
\includegraphics[height=100mm]{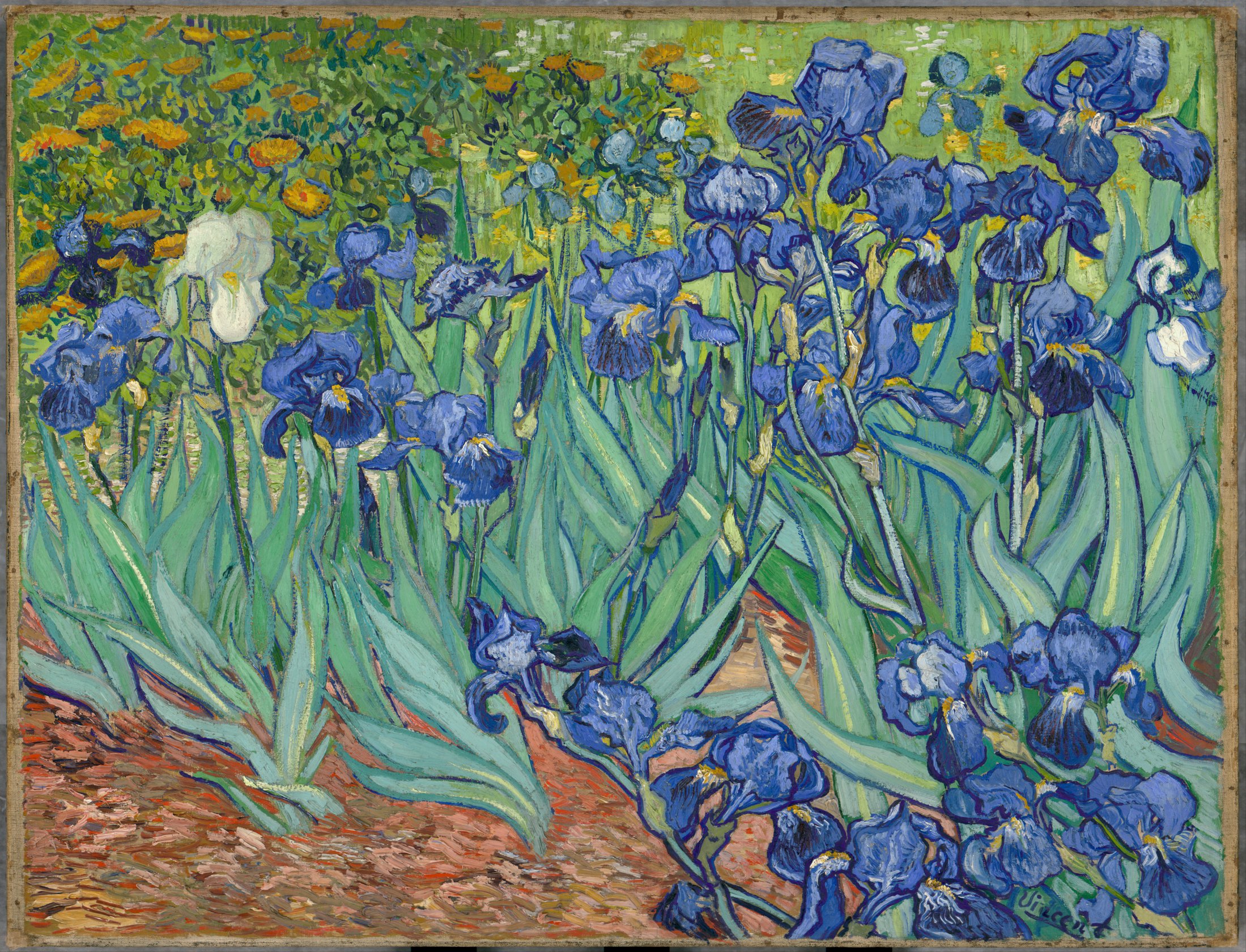}\\
\includegraphics[height=100mm]{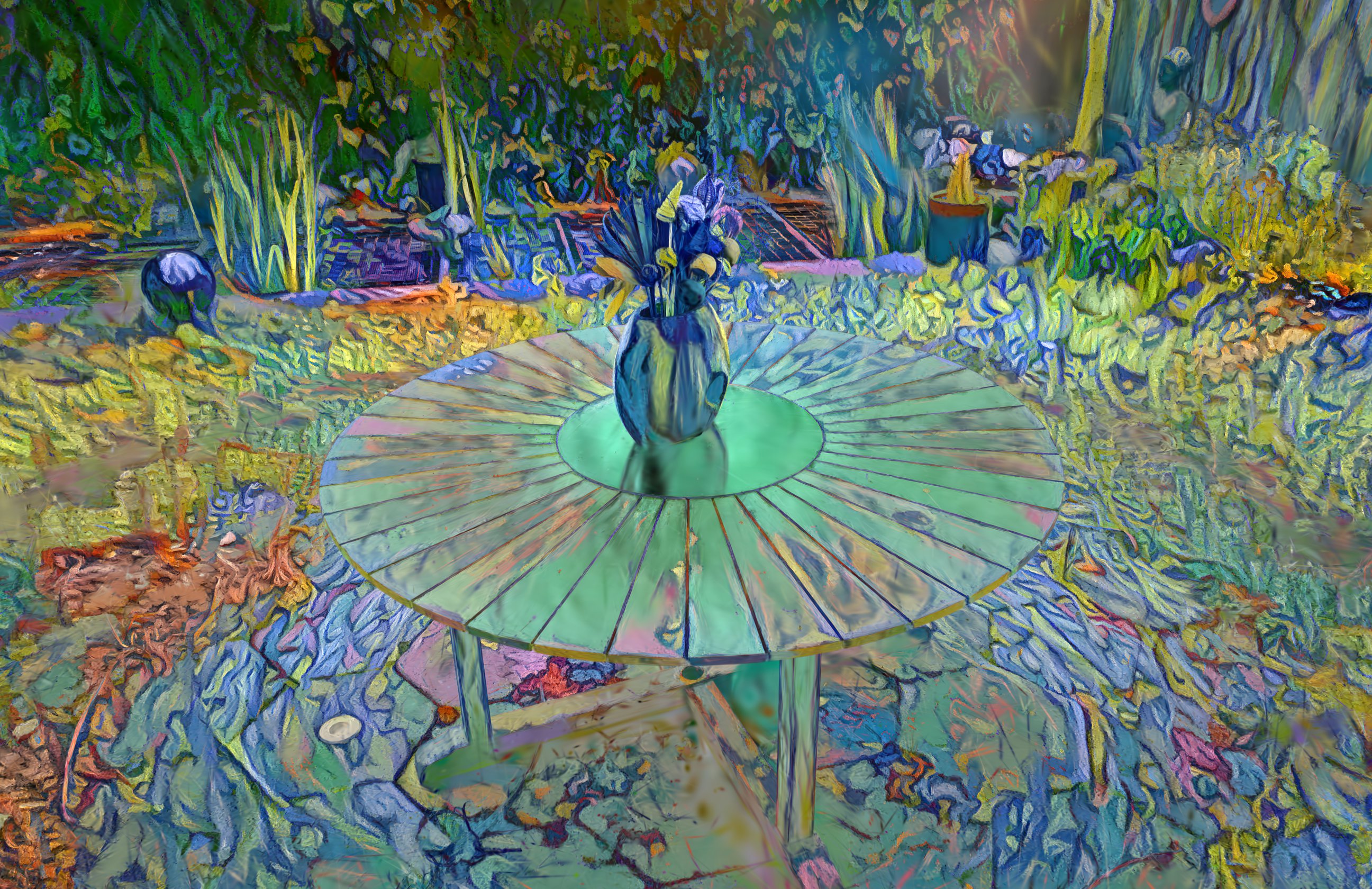}
\end{center}
\caption{Full view display of the example 3/8 of the teaser figure with the style image (size 4398$\times$3361) displayed at the same scale as the rendered image (size 5187$\times$3361). 
Images have been downscaled by a factor 2 and compressed using jpeg.}
\label{fig:supp_mat_teaser_full_3}
\end{figure*}

\begin{figure*}
\begin{center}
\includegraphics[height=87.2656947337102mm]{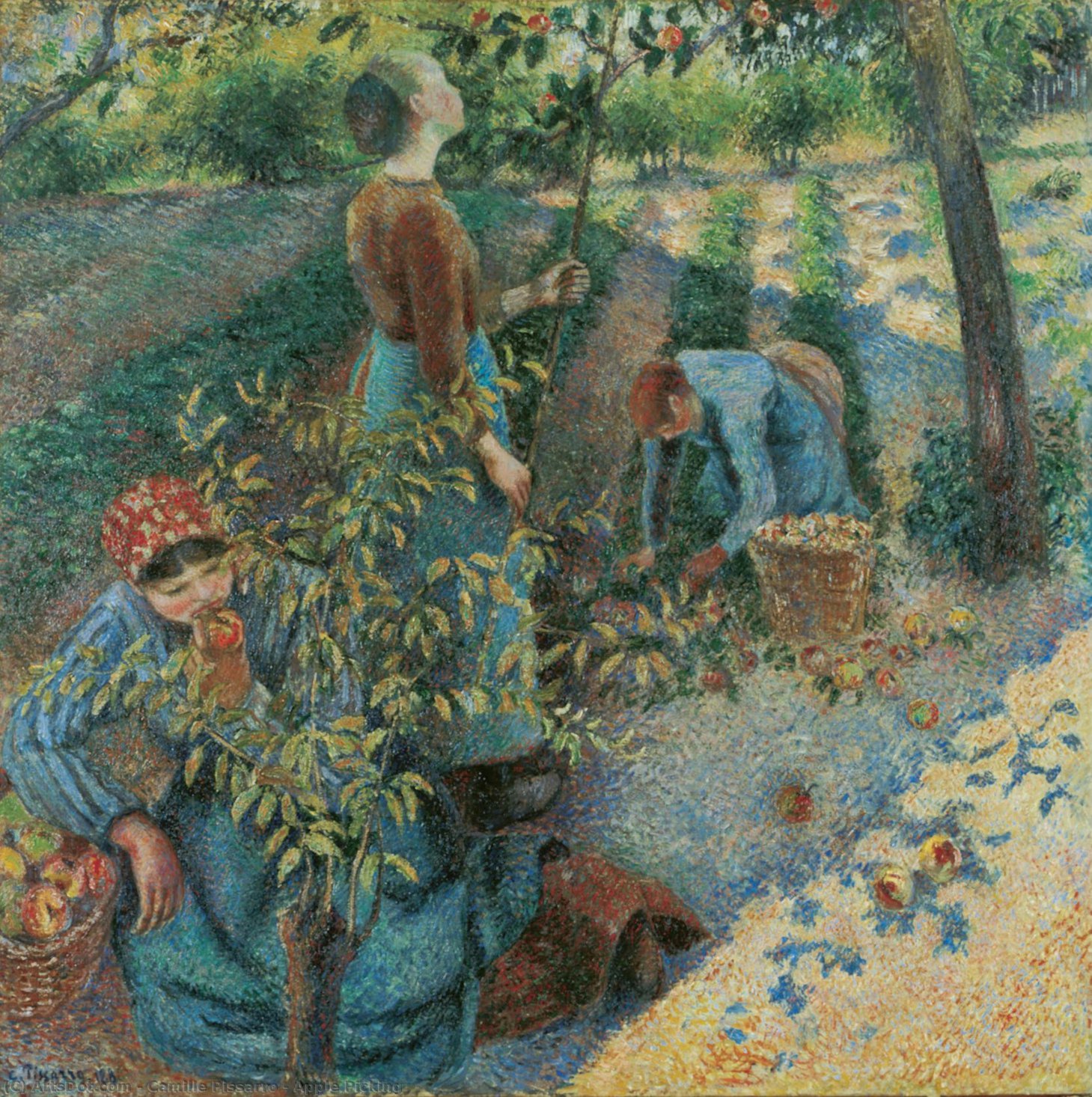}\\
\includegraphics[height=100mm]{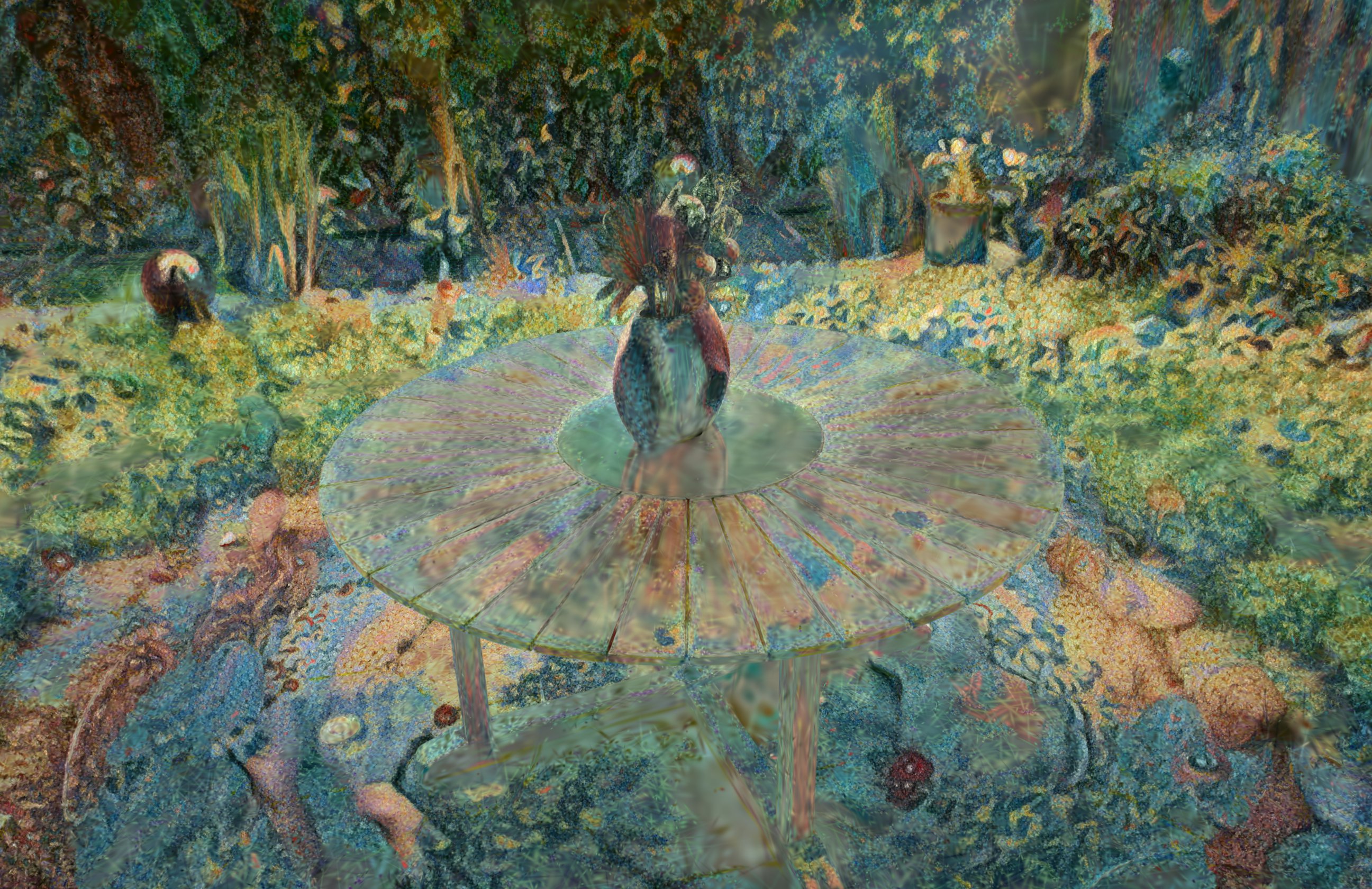}
\end{center}
\caption{Full view display of the example 4/8 of the teaser figure with the style image (size 4398$\times$3361) displayed at the same scale as the rendered image (size 5187$\times$3361). 
Images have been downscaled by a factor 2 and compressed using jpeg.}
\label{fig:supp_mat_teaser_full_4}
\end{figure*}

\begin{figure*}
\begin{center}
\includegraphics[height=100mm]{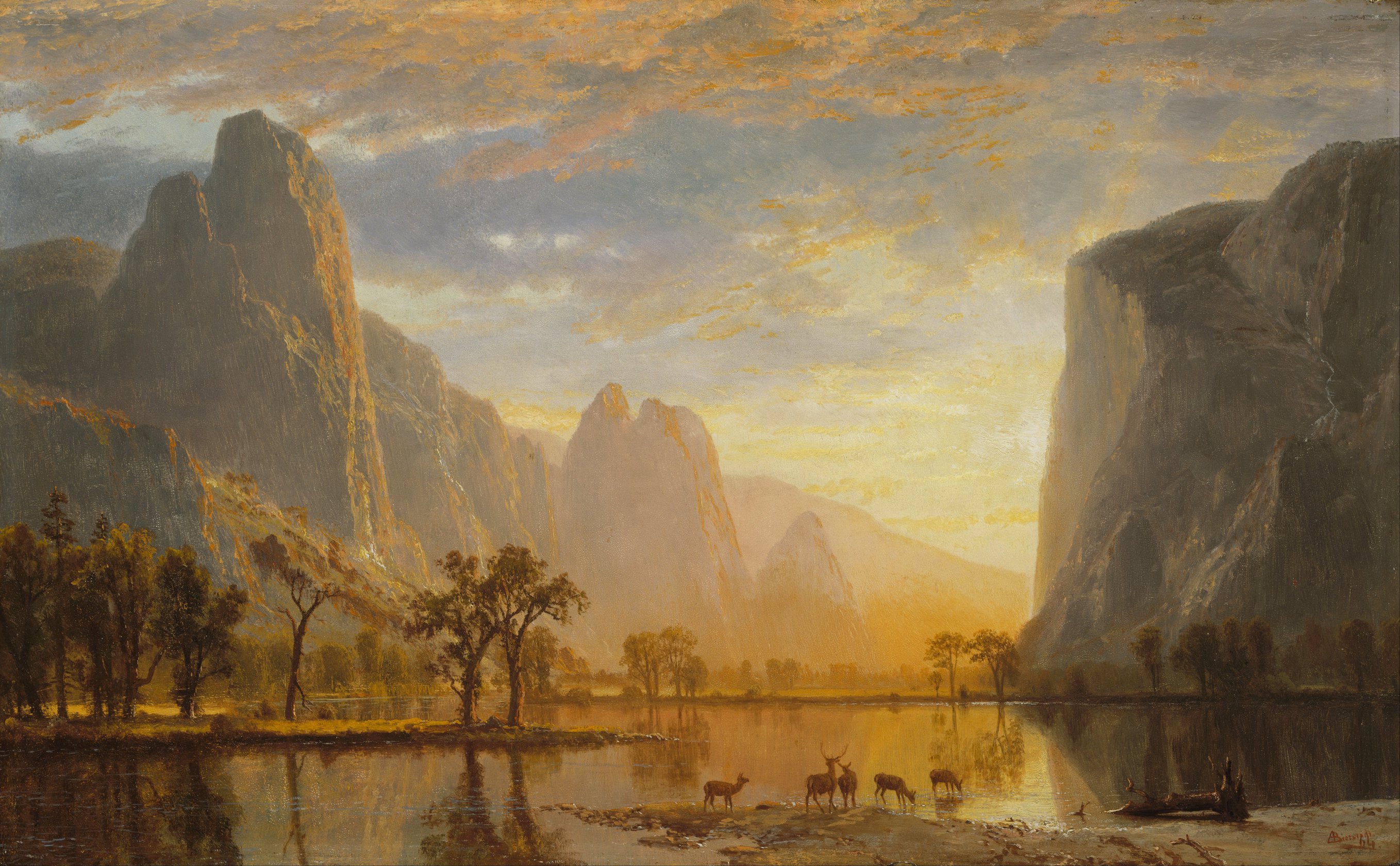}\\
\includegraphics[height=100mm]{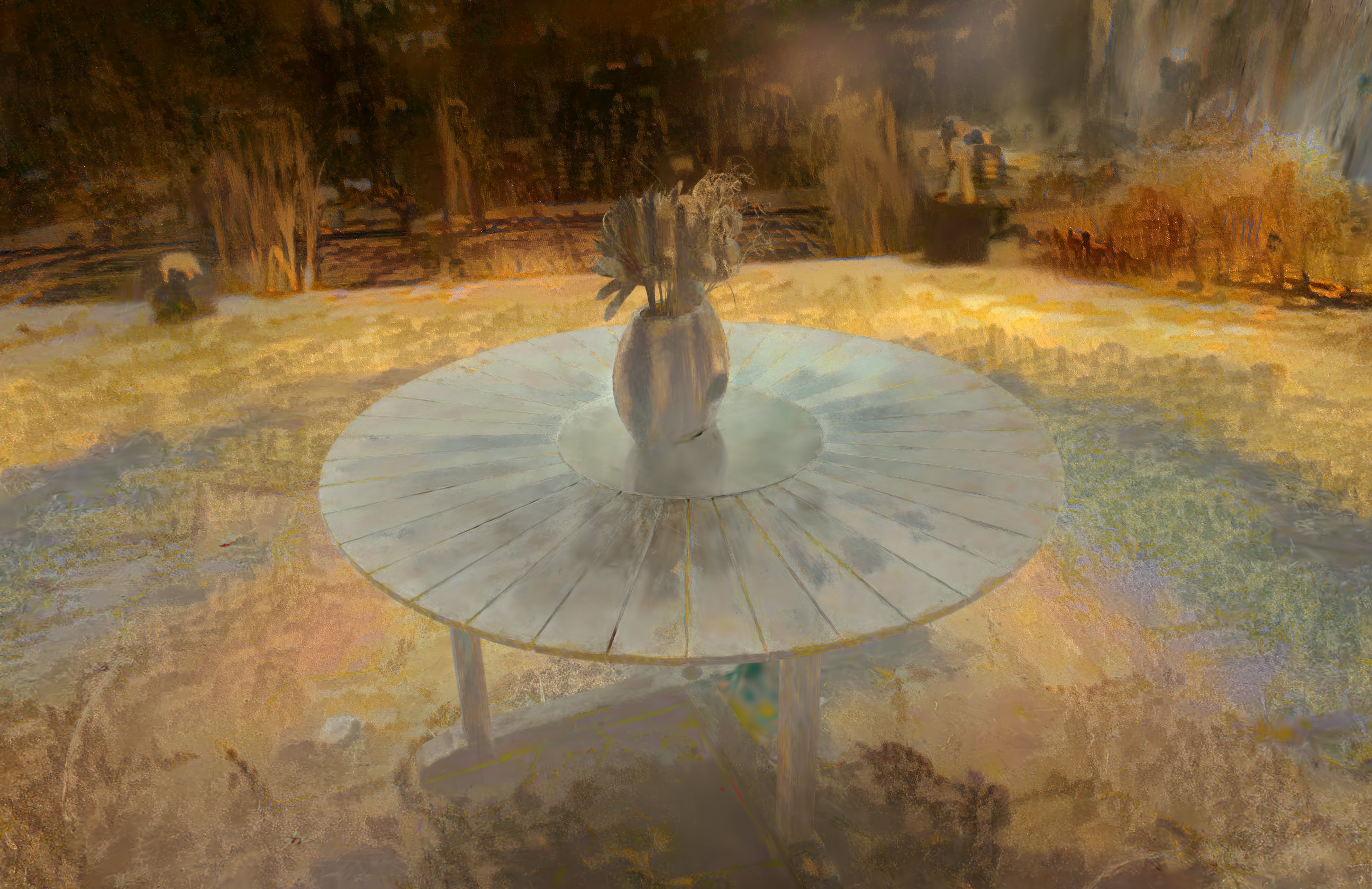}
\end{center}
\caption{Full view display of the example 5/8 of the teaser figure with the style image (size 5433$\times$3361) displayed at the same scale as the rendered image (size 5187$\times$3361). 
Images have been downscaled by a factor 2 and compressed using jpeg.}
\label{fig:supp_mat_teaser_full_5}
\end{figure*}

\begin{figure*}
\begin{center}
\includegraphics[height=20.58911038381434mm]{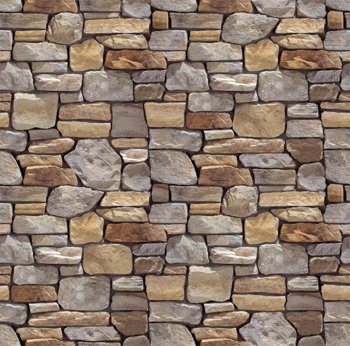}\\
\includegraphics[height=100mm]{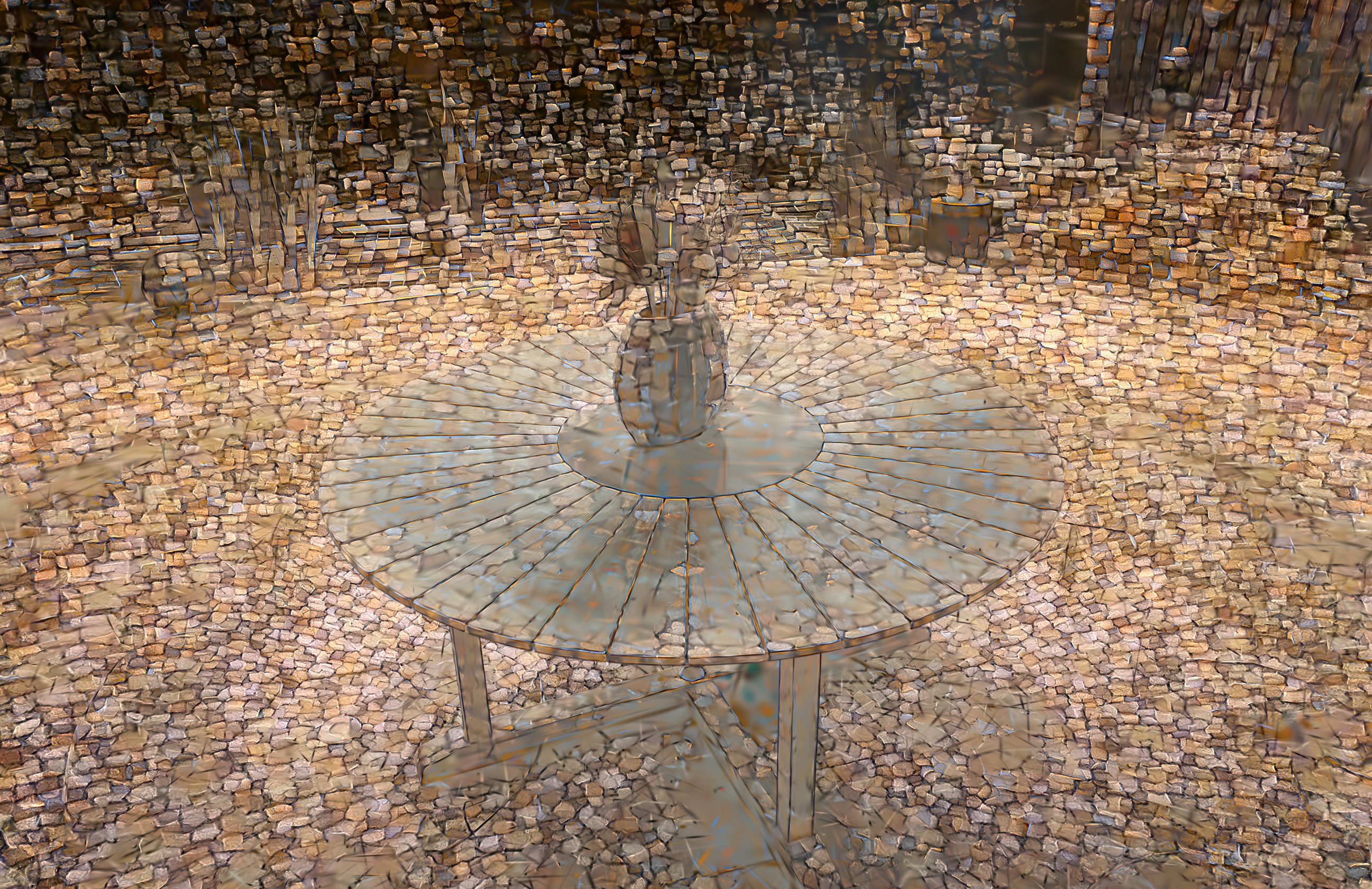}
\end{center}
\caption{Full view display of the example 6/8 of the teaser figure with the style image (size 700$\times$692) displayed at the same scale as the rendered image (size 5187$\times$3361). 
Images have been downscaled by a factor 2 and compressed using jpeg.}
\label{fig:supp_mat_teaser_full_6}
\end{figure*}

\begin{figure*}
\begin{center}
\includegraphics[height=23.237131806010115mm]{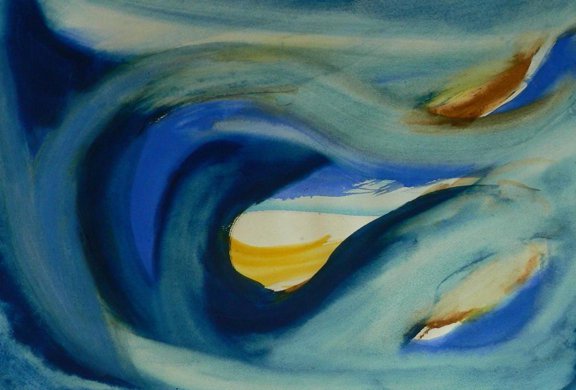}\\
\includegraphics[height=100mm]{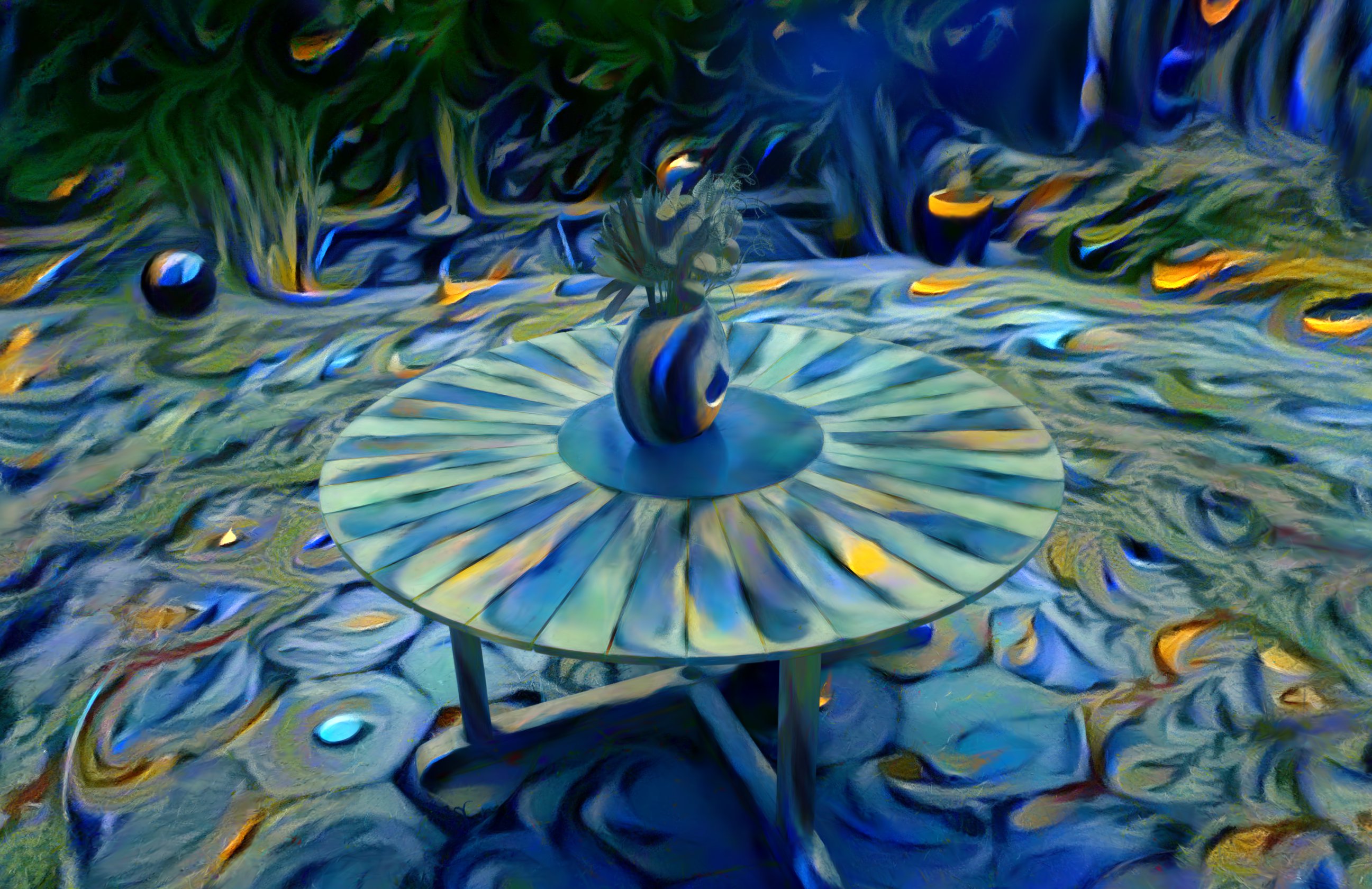}
\end{center}
\caption{Full view display of the example 7/8 of the teaser figure with the style image (size 1152$\times$781) displayed at the same scale as the rendered image (size 5187$\times$3361). 
Images have been downscaled by a factor 2 and compressed using jpeg.}
\label{fig:supp_mat_teaser_full_7}
\end{figure*}

\begin{figure*}
\begin{center}
\includegraphics[height=30.46712288009521mm]{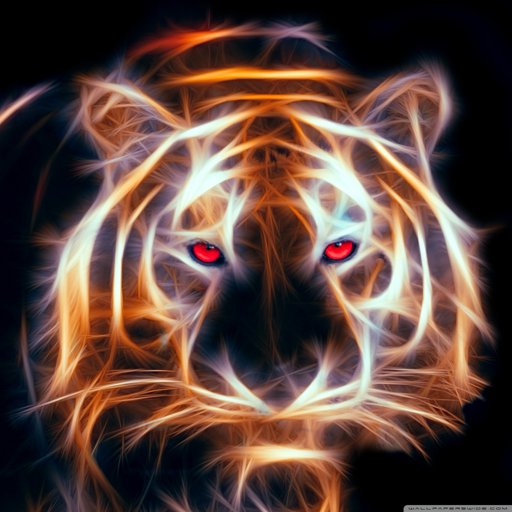}\\
\includegraphics[height=100mm]{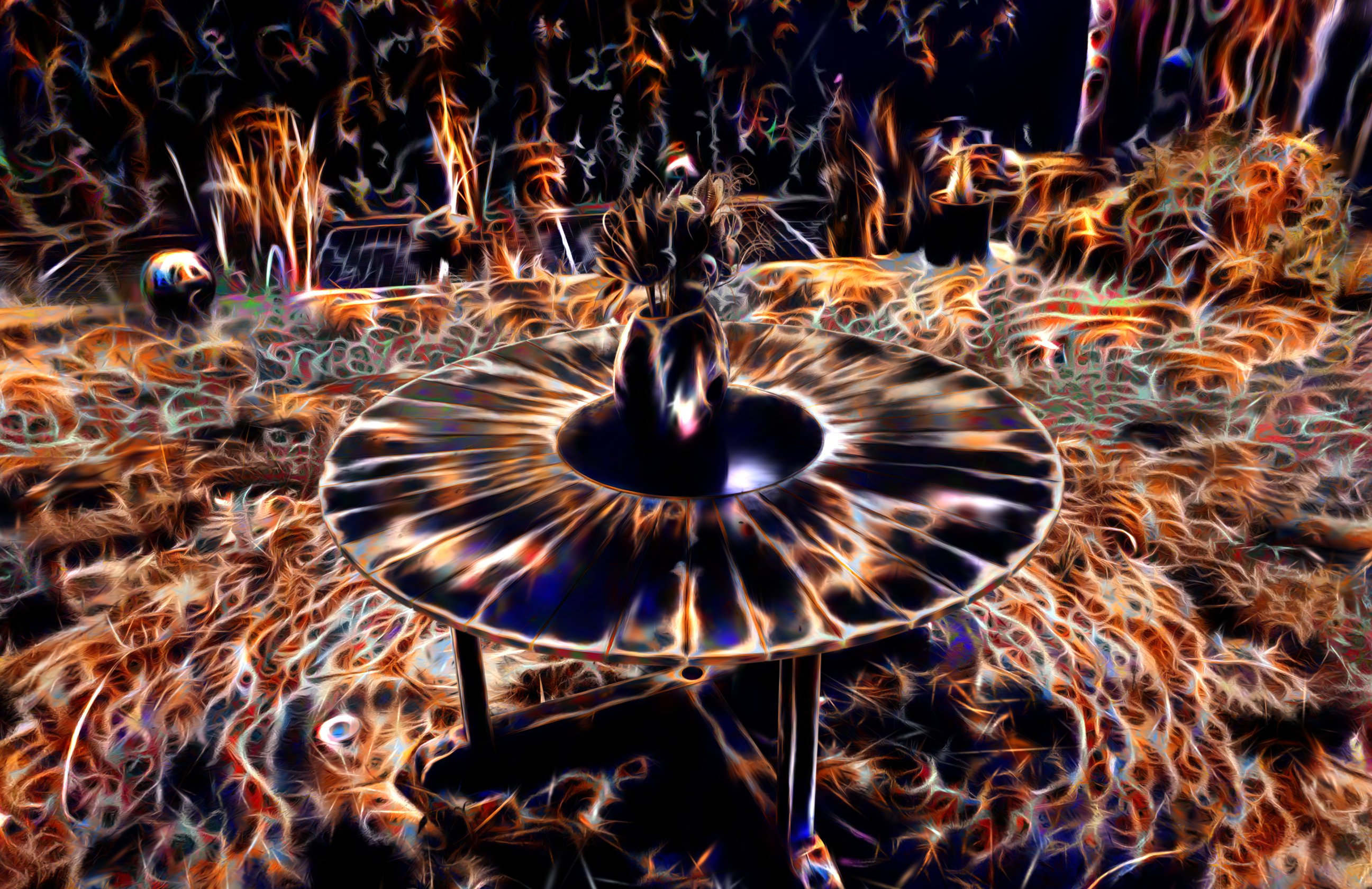}
\end{center}
\caption{Full view display of the example 8/8 of the teaser figure with the style image (size 1024$\times$1024) displayed at the same scale as the rendered image (size 5187$\times$3361). 
Images have been downscaled by a factor 2 and compressed using jpeg.}
\label{fig:supp_mat_teaser_full_8}
\end{figure*}

Due to space limitation, style images of the main paper's teaser figure have been displayed as tiny images regardless of their resolution. 
Figures~\ref{fig:supp_mat_teaser_full_1} to~\ref{fig:supp_mat_teaser_full_8} show the eight pairs of images of this figure in full size to better appreciate the multiscale details of the style images and their corresponding stylized results.
Each style image is displayed at the same resolution as the rendered view so that one can observe that the style features are reproduced with the same size (see e.g. the stone wall of Figure~\ref{fig:supp_mat_teaser_full_6}).

\section{Comparison experiments}
\label{sec:comparison_experiments}

As said in the main paper, we performed a thorough comparative study using 40 3D style transfer experiments using 9 different scenes from previous works~\cite{Barron_etal_Mip-NeRF360_unbounded_antialiased_neural_radiance_fields_CVPR2022, Fridovich-Keil_Yu_etal_plenoxels_radiance_fields_without_neural_networks_CVPR2022, Kerbl_etal_3D_Gaussian_splatting_for_real-time_radiance_field_rendering_SIGGRAPH2023} and various style images.
We compare our results with the NeRF-based ARF~\cite{Zhang_etal_arf_artistic_radiance_fields_ECCV2022} and 
the 3DGS-based StyleGaussian~\cite{Liu_etal_StyleGaussian_instant_3D_style_transfer_with_Gaussian_splatting_ArXiv2024} algorithms using their public implementations\footnote{\url{https://github.com/Kai-46/ARF-svox2}; \url{https://github.com/Kunhao-Liu/StyleGaussian}}.

\begin{figure*}
\includegraphics[width=\linewidth]{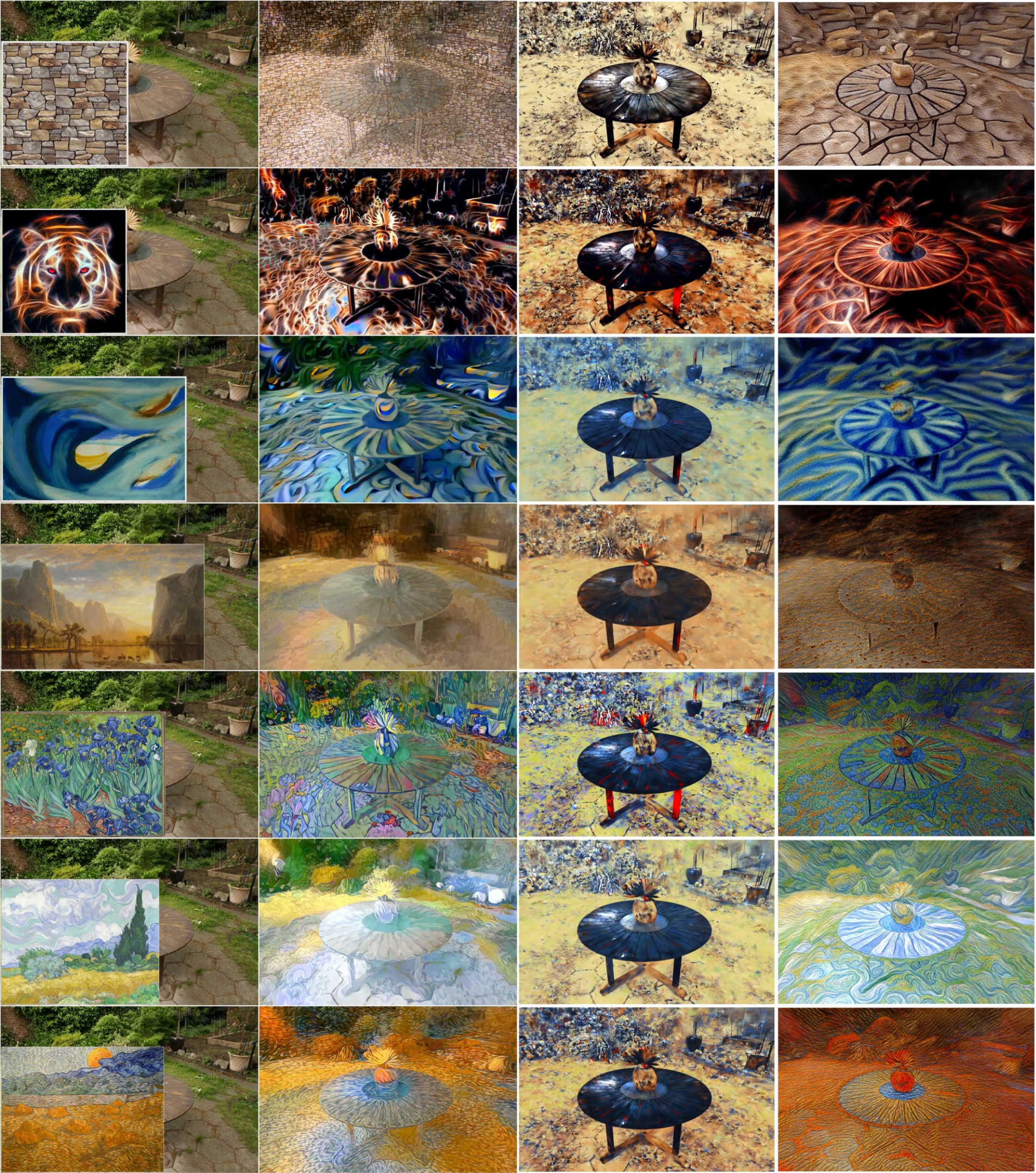}
\caption{Comparative experiments using the garden scene. From left to right: Content and style, SGSST (ours), StyleGaussian, ARF. Content image size is 5187$\times$3361.}
\label{fig:comparison_garden}
\end{figure*}

\begin{figure*}
\includegraphics[width=\linewidth]{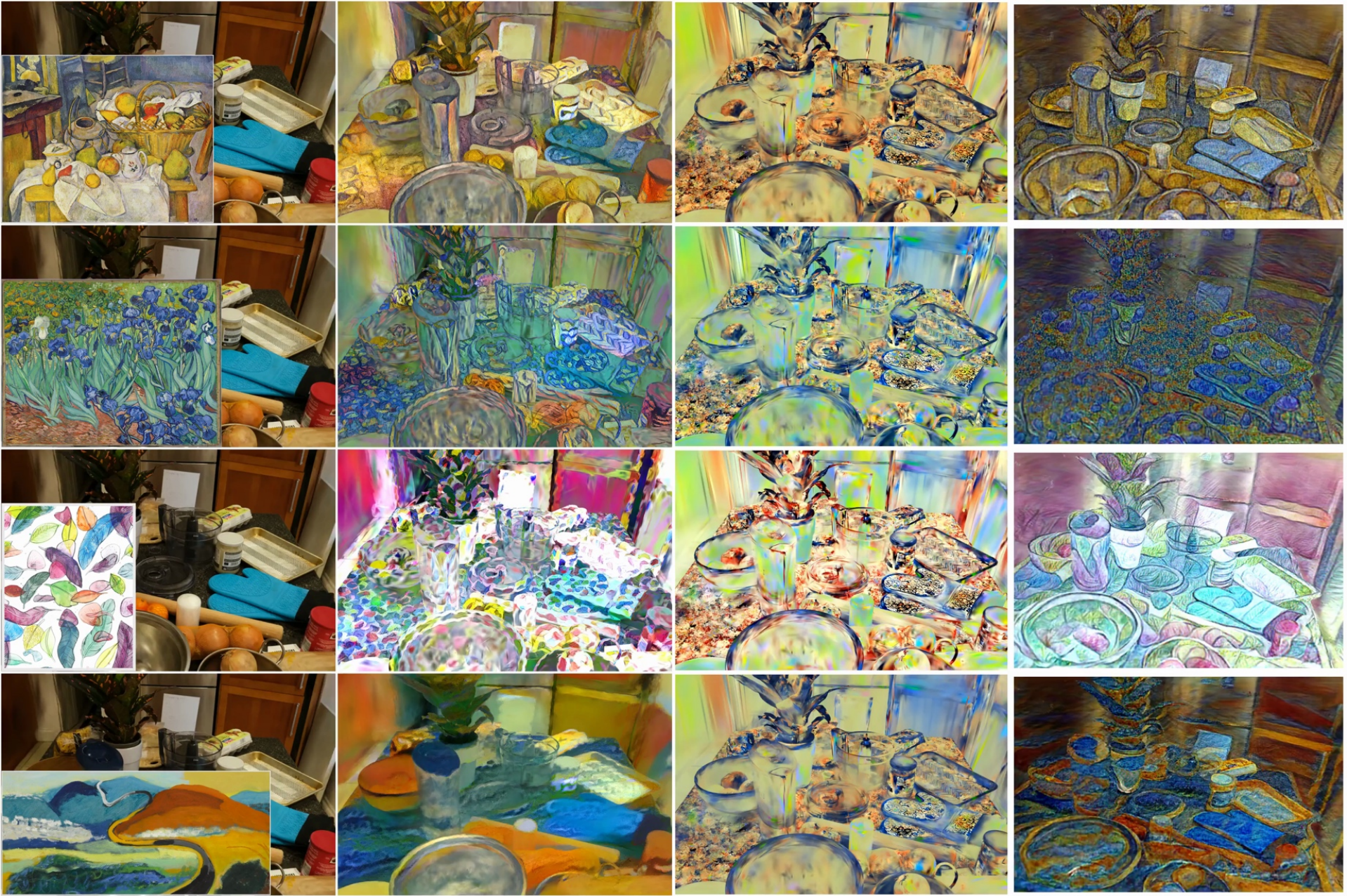}
\caption{Comparative experiments using the counter scene. From left to right: Content and style, SGSST (ours), StyleGaussian, ARF. Content image size is 3115$\times$2076.}
\label{fig:comparison_counter}
\end{figure*}

\begin{figure*}
\includegraphics[width=\linewidth]{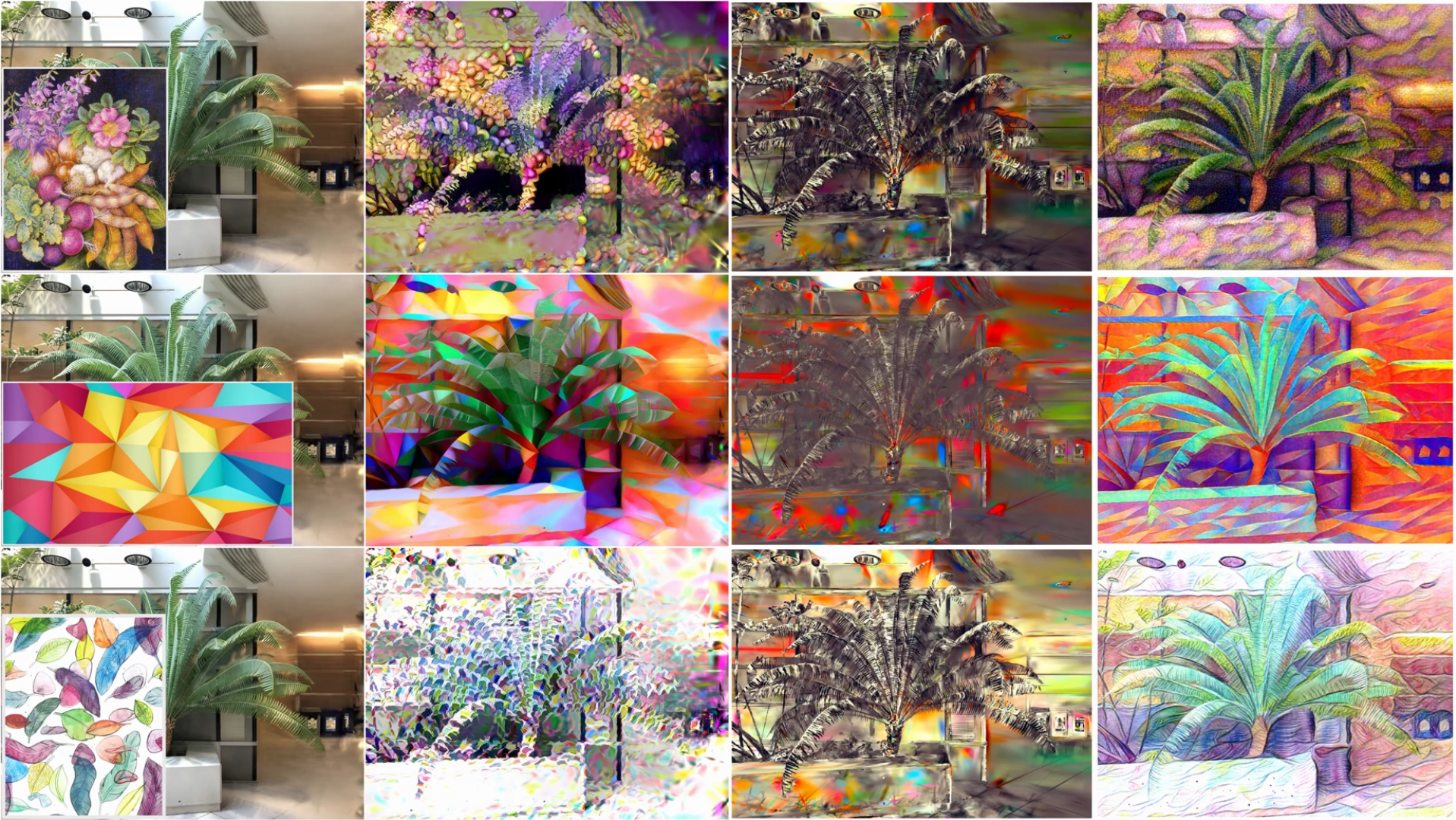}
\caption{Comparative experiments using the fern scene. From left to right: Content and style, SGSST (ours), StyleGaussian, ARF. Content image size is 4032$\times$3024.}
\label{fig:comparison_fern}
\end{figure*}

\begin{figure*}
\includegraphics[width=\linewidth]{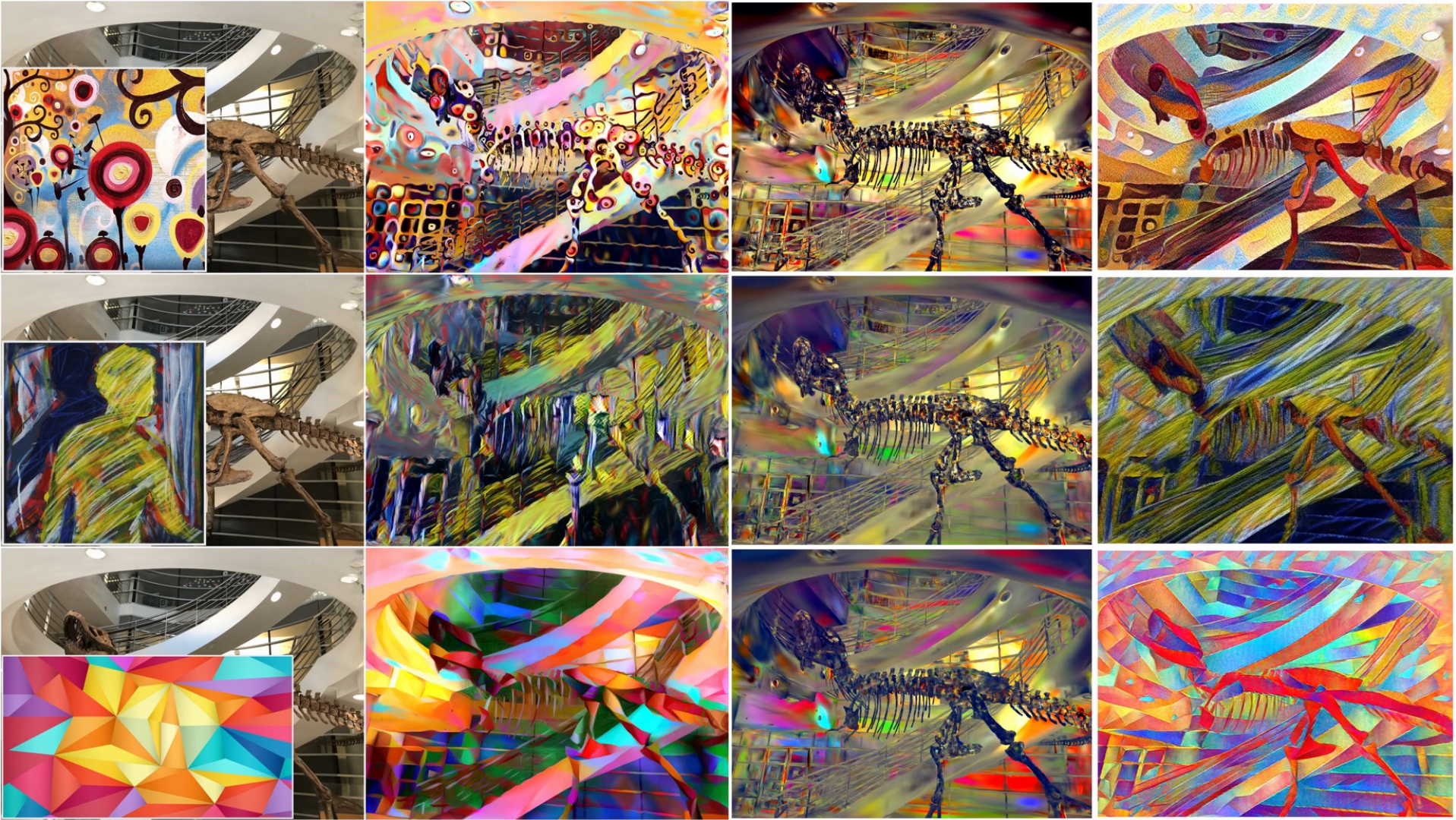}
\caption{Comparative experiments on the t-rex scene. From left to right: Content and style, SGSST (ours), StyleGaussian, ARF. Content image size is 4032$\times$3024.}
\label{fig:comparison_trex}
\end{figure*}

\begin{figure*}
\includegraphics[width=\linewidth]{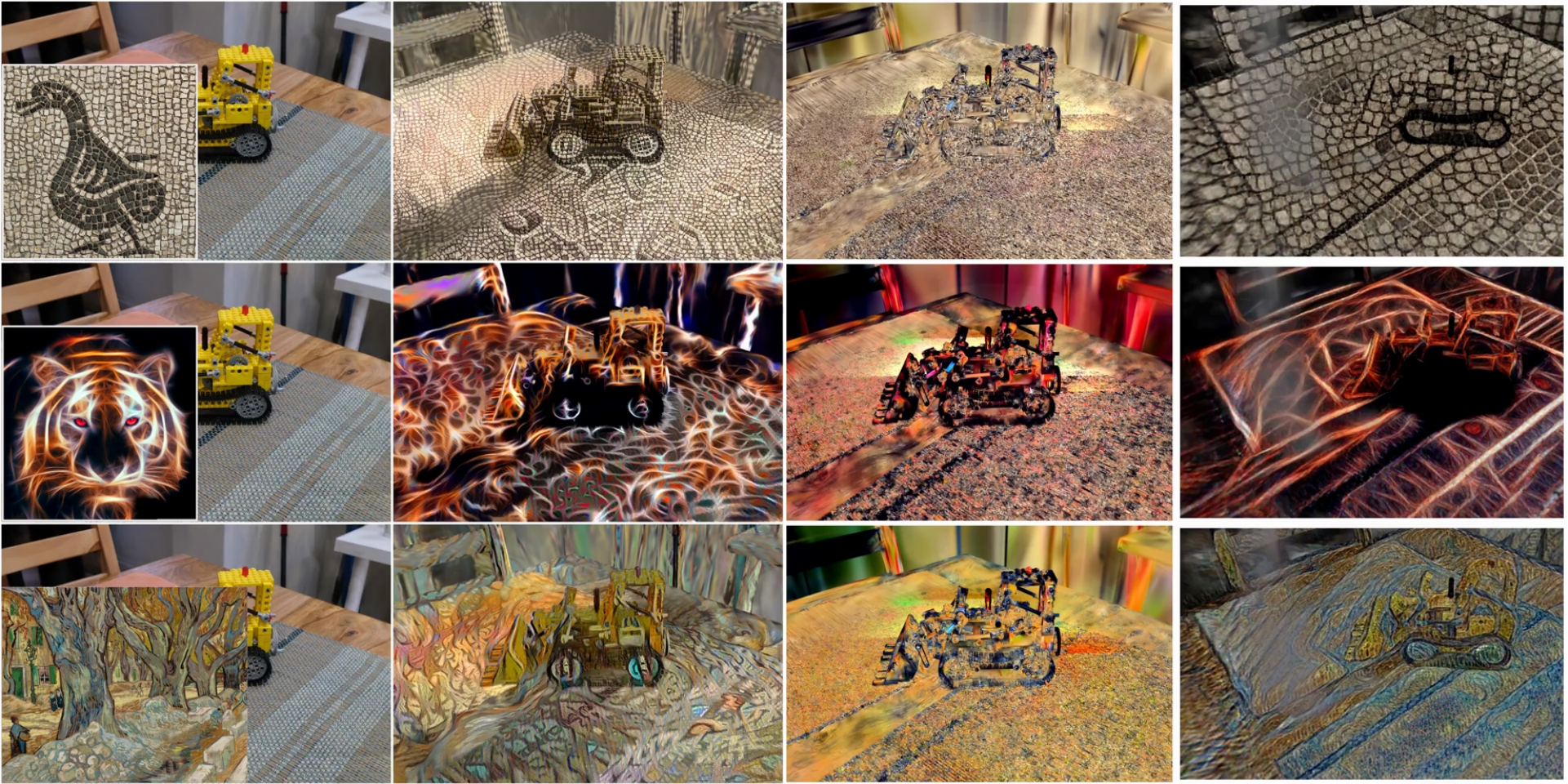}
\caption{Comparative experiments on the kitchen scene. From left to right: Content and style, SGSST (ours), StyleGaussian, ARF. Content image size is 3115$\times$2078.}
\label{fig:comparison_kitchen}
\end{figure*}

\begin{figure*}
\includegraphics[width=\linewidth]{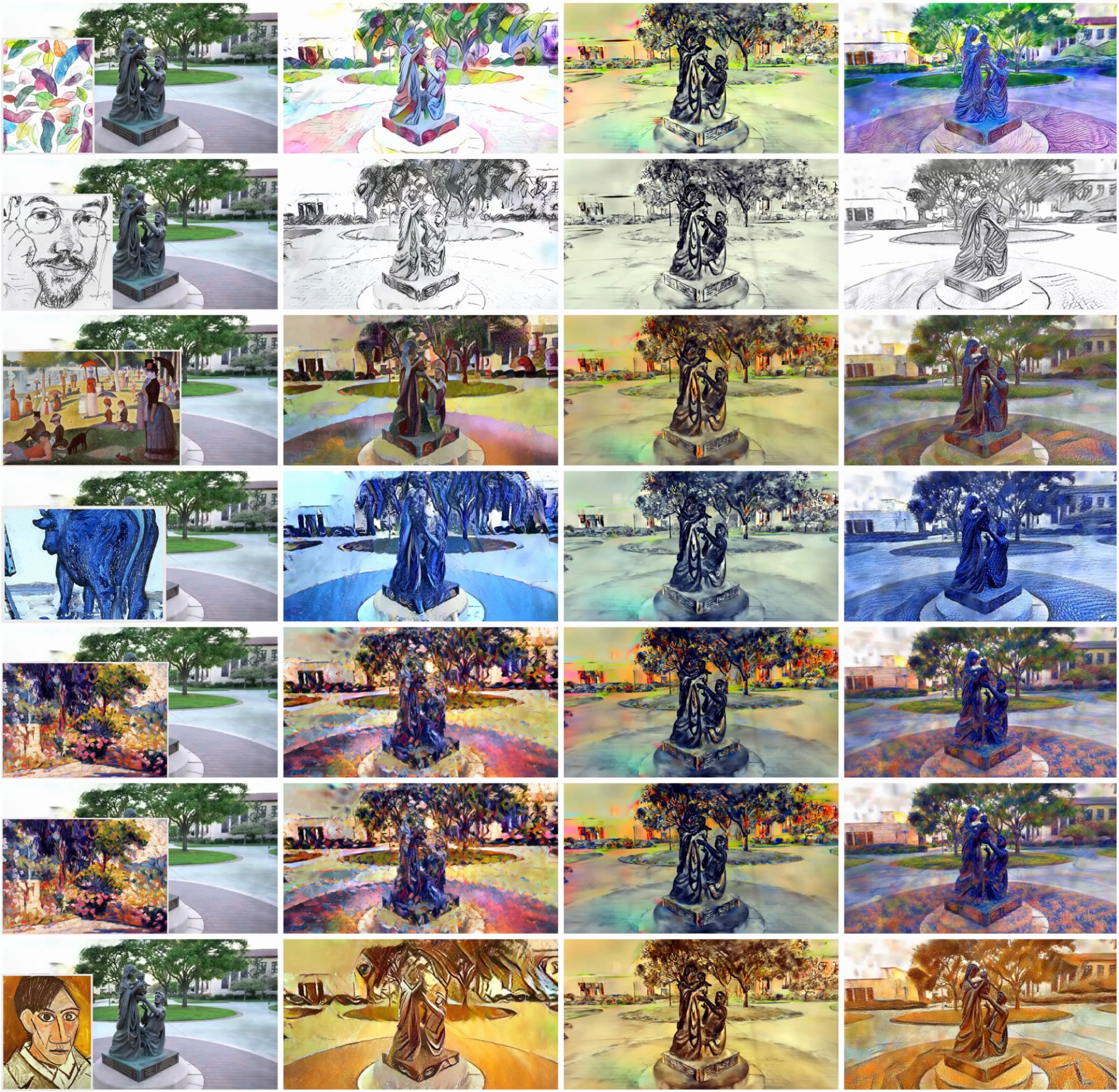}
\caption{Comparative experiments on the family scene. From left to right: Content and style, SGSST (ours), StyleGaussian, ARF. Content image size is 977$\times$544.}
\label{fig:comparison_family}
\end{figure*}

\begin{figure*}
\includegraphics[width=\linewidth]{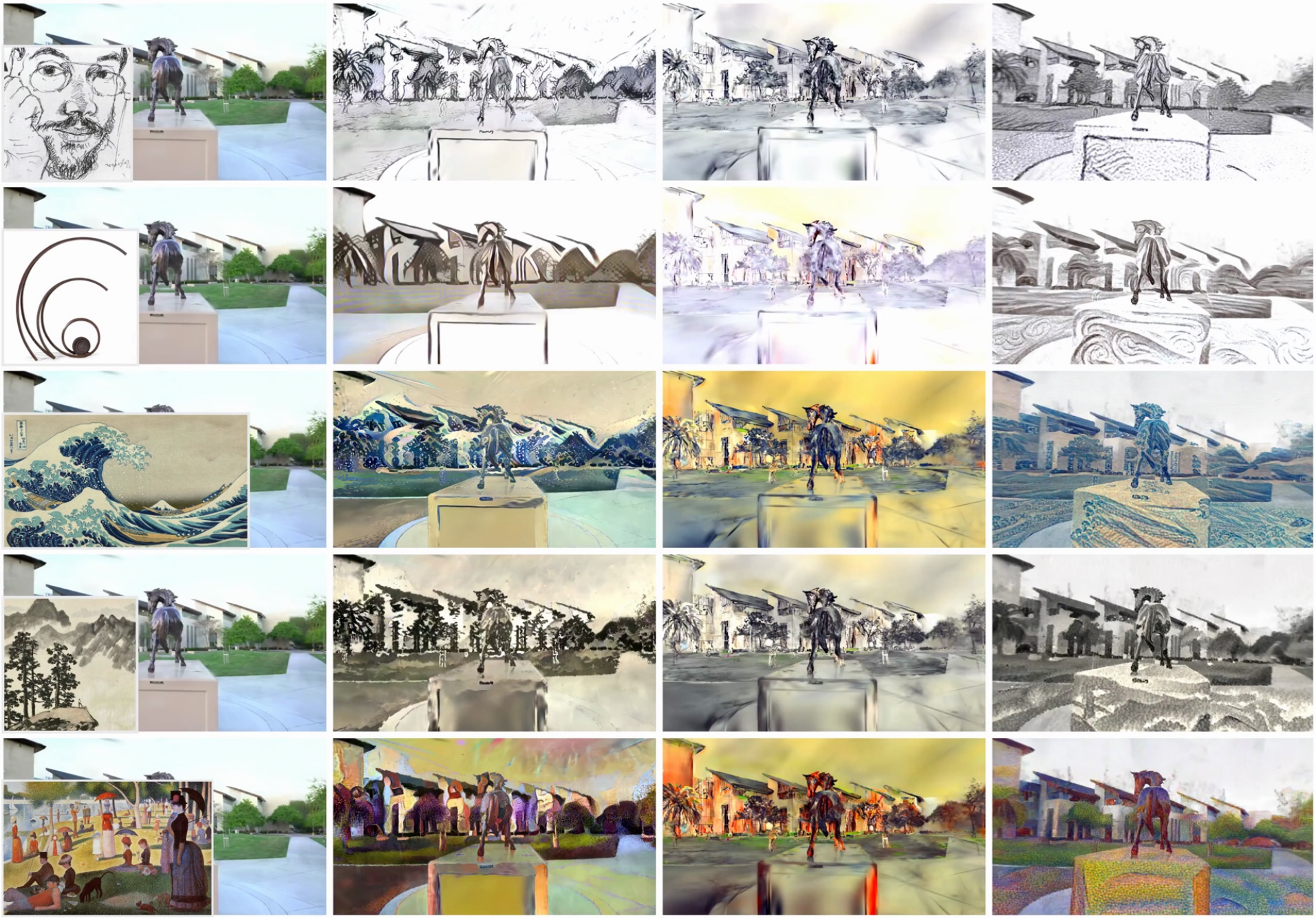}
\caption{Comparative experiments on the horse scene. From left to right: Content and style, SGSST (ours), StyleGaussian, ARF. Content image size is 976$\times$544.}
\label{fig:comparison_horse}
\end{figure*}

\begin{figure*}
\includegraphics[width=\linewidth]{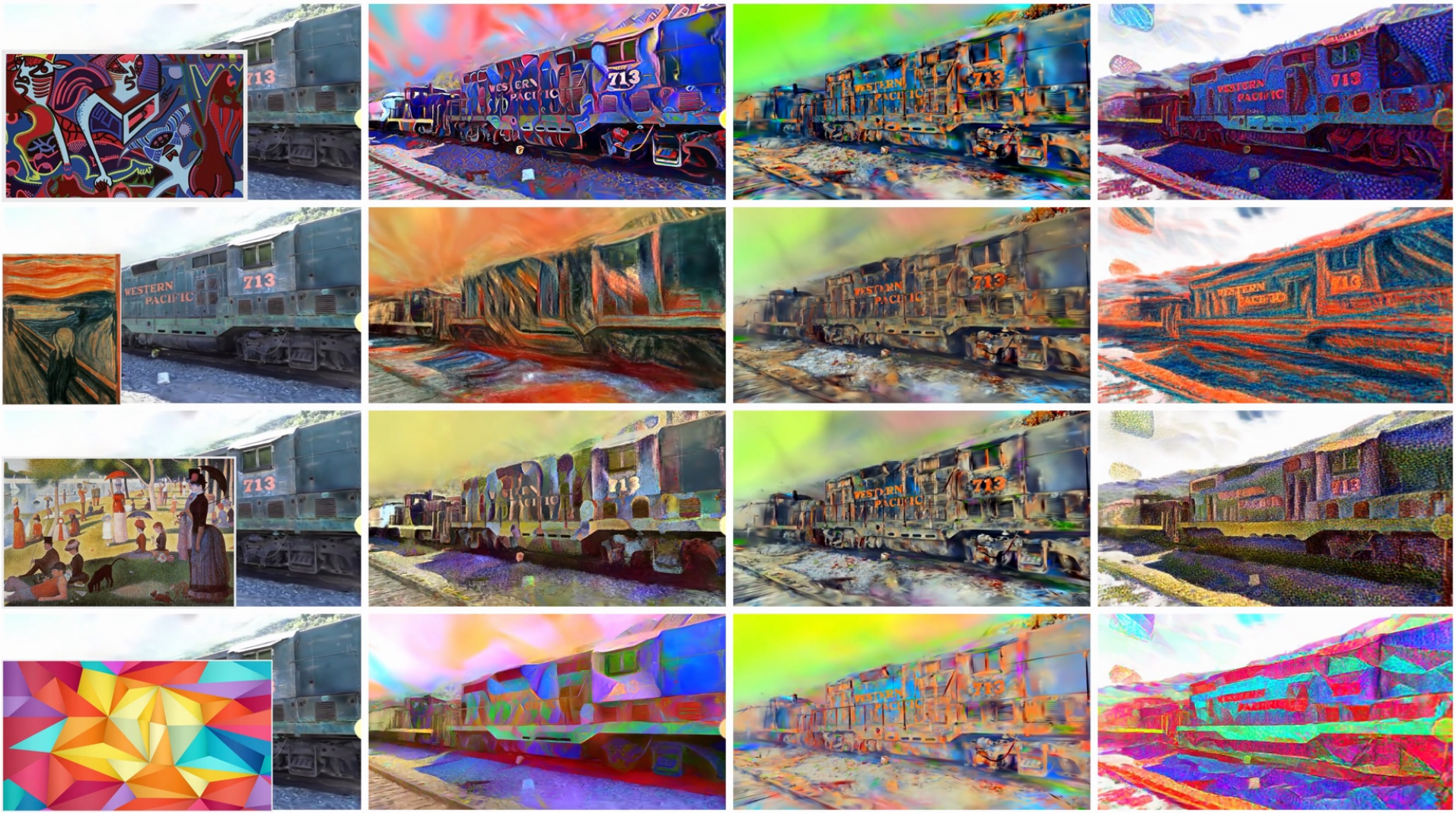}
\caption{Comparative experiments on the train scene. From left to right: Content and style, SGSST (ours), StyleGaussian, ARF. Content image size is 980$\times$545.}
\label{fig:comparison_train}
\end{figure*}

\begin{figure*}
\includegraphics[width=\linewidth]{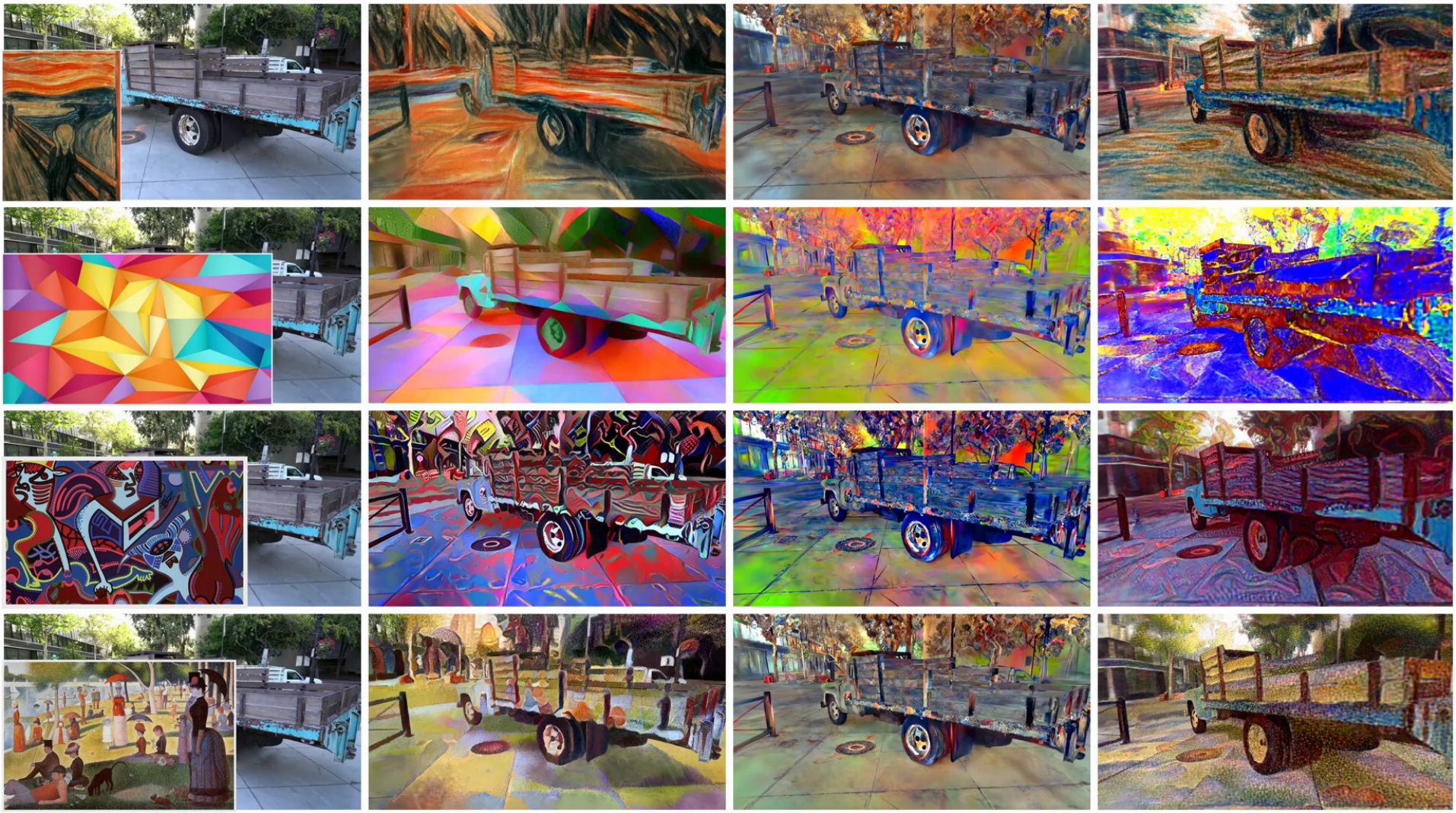}
\caption{Comparative experiments on the truck scene. From left to right: Content and style, SGSST (ours), StyleGaussian, ARF. Content image size is 979$\times$546.}
\label{fig:comparison_truck}
\end{figure*}

Figures~\ref{fig:comparison_garden} to \ref{fig:comparison_truck} display a rendered view for each of these 40 experiments.
Let us recall that for the HR scenes (Figures~\ref{fig:comparison_garden} to \ref{fig:comparison_kitchen}) our approach is the only one working at high-resolution.
While SGSST produces outputs having the content size, StyleGaussian outputs are limited in resolution to a maximal width of 1600 or maximal height of 1200, 
and for ARF the content images have been downscaled by a factor 4 to obtain a low-resolution input suitable for ARF (see Section~\ref{sec:arf_hr} below).
Video versions of these figures are available at:
\url{https://www.idpoisson.fr/galerne/sgsst/comparison_web.html}.

\begin{figure*}
\includegraphics[width=\textwidth]{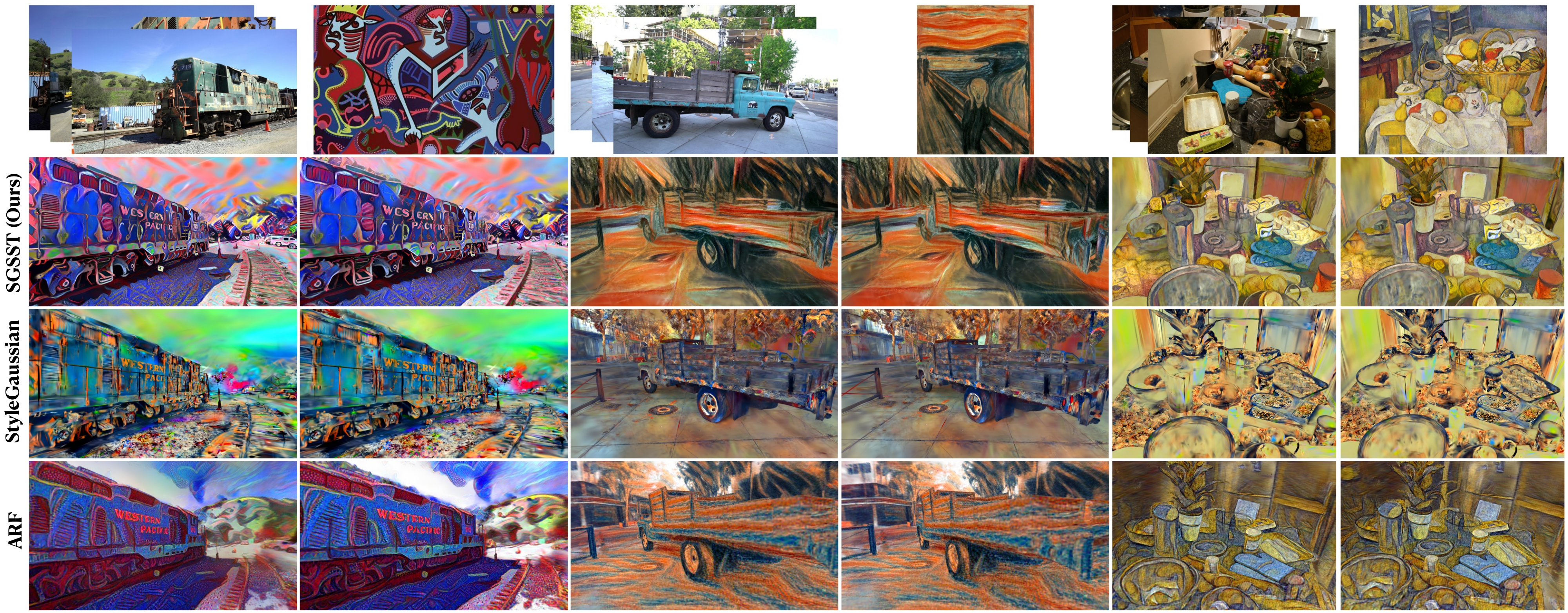}
\caption{\textbf{Comparison of SGSST (ours, top) with StyleGaussian~\cite{Liu_etal_StyleGaussian_instant_3D_style_transfer_with_Gaussian_splatting_ArXiv2024} (middle) and ARF~\cite{Zhang_etal_arf_artistic_radiance_fields_ECCV2022} (bottom) with short range views.}
 From left to right the content resolutions are 
980$\times$545 (train), 
979$\times$546 (truck), 
and 
3115$\times$2076 (counter).
For the first two  examples, the various outputs keep the resolution of the content, but for the HR counter scene, the output sizes are 3115$\times$2076 for SGSST, 1600$\times$1066 for StyleGaussian and 779$\times$519 for ARF (see supp. mat. for ARF results without downscaling).
Thanks to its multiscale global VGG statistics, SGSST is the most faithful method regarding style consistency.%
} 
\label{fig:comparison_short_range}
\end{figure*}

Moreover, we provide with Figure~\ref{fig:comparison_short_range} a second version of the comparison figure (Figure 4) with close views to highlight the texture consistency of each method.

\section{ARF and high resolution inputs}
\label{sec:arf_hr}

\begin{figure*}
\includegraphics[width=\linewidth]{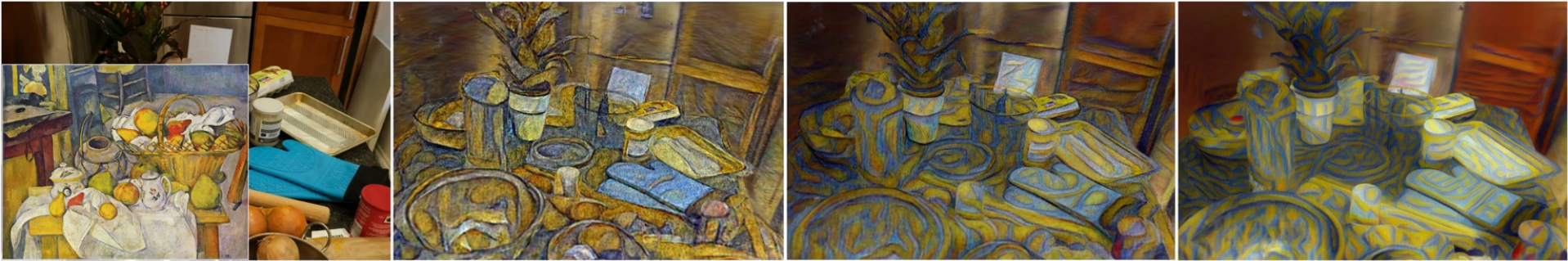}
\caption{ARF outputs for the HR style transfer: example of the main paper with various downscaling factors. ARF produces good stylization results  for inputs of moderate resolution only. From left to right: Scene and style (input size is 3115$\times$2076), 
ARF result with input downscaled by 4 (size 779$\times$519), 
ARF result with input downscaled by 2 (size 1557$\times$1038), 
ARF result with original HR (size 3115$\times$2076).
}
\label{fig:arf_scale}
\end{figure*}

ARF~\cite{Zhang_etal_arf_artistic_radiance_fields_ECCV2022} uses Nearest Neighbor Feature Matching (NNFM) of a single layer of VGG features for fine tuning a plenoxel radiance field~\cite{Fridovich-Keil_Yu_etal_plenoxels_radiance_fields_without_neural_networks_CVPR2022}.
It produces high-quality results at moderate resolution.
While comparing our results with ARF, we observed that this algorithm does not produce visually satisfying results for high-resolution scenes. 
This is illustrated by Figure~\ref{fig:arf_scale} where one can observe that the style transfer quality decreases as the input size increases.
To allow a fair comparison we decided to downscale images by a factor 4 for the high-resolution scene as a preprocess for ARF.

Although it has been shown that NNFM is superior to Gram feature matrix optimization for NeRF style transfer when optimizing  for a single VGG layer~\cite{Zhang_etal_arf_artistic_radiance_fields_ECCV2022},
our results show that optimizing for a (slightly corrected~\cite{Galerne_etal_scaling_painting_style_transfer_EGSR2024}) Gram-based loss using several image scales and five VGG layers for each scale is an effective solution for applying high quality style transfer at UHR.

\section{Details on the perceptual study}
\label{sec:perceptual_study}

As described in the main paper, a comparative perceptual study was conducted using the 40 3D style transfer experiments presented in Section~\ref{sec:comparison_experiments} (Figures~\ref{fig:comparison_garden} to \ref{fig:comparison_truck}).
For each experiment, they were asked to pick the image that appeared to be the most faithful to the style image among the three displayed results.
Each participant was shown ten random experiments and participation was voluntary.
Figure~\ref{fig:arf_scale} is an example of such an experiment displaying the style image (top left image) and three views of the scenes stylized by SGSST, ARF and StyleGaussian respectively and displayed in random order. 
To choose between one of the three results, the participant had to press the left arrow key to select the bottom left result, the up arrow key to select the top right result and the down arrow key to select the bottom right result.

\begin{figure*}
\includegraphics[width=\linewidth]{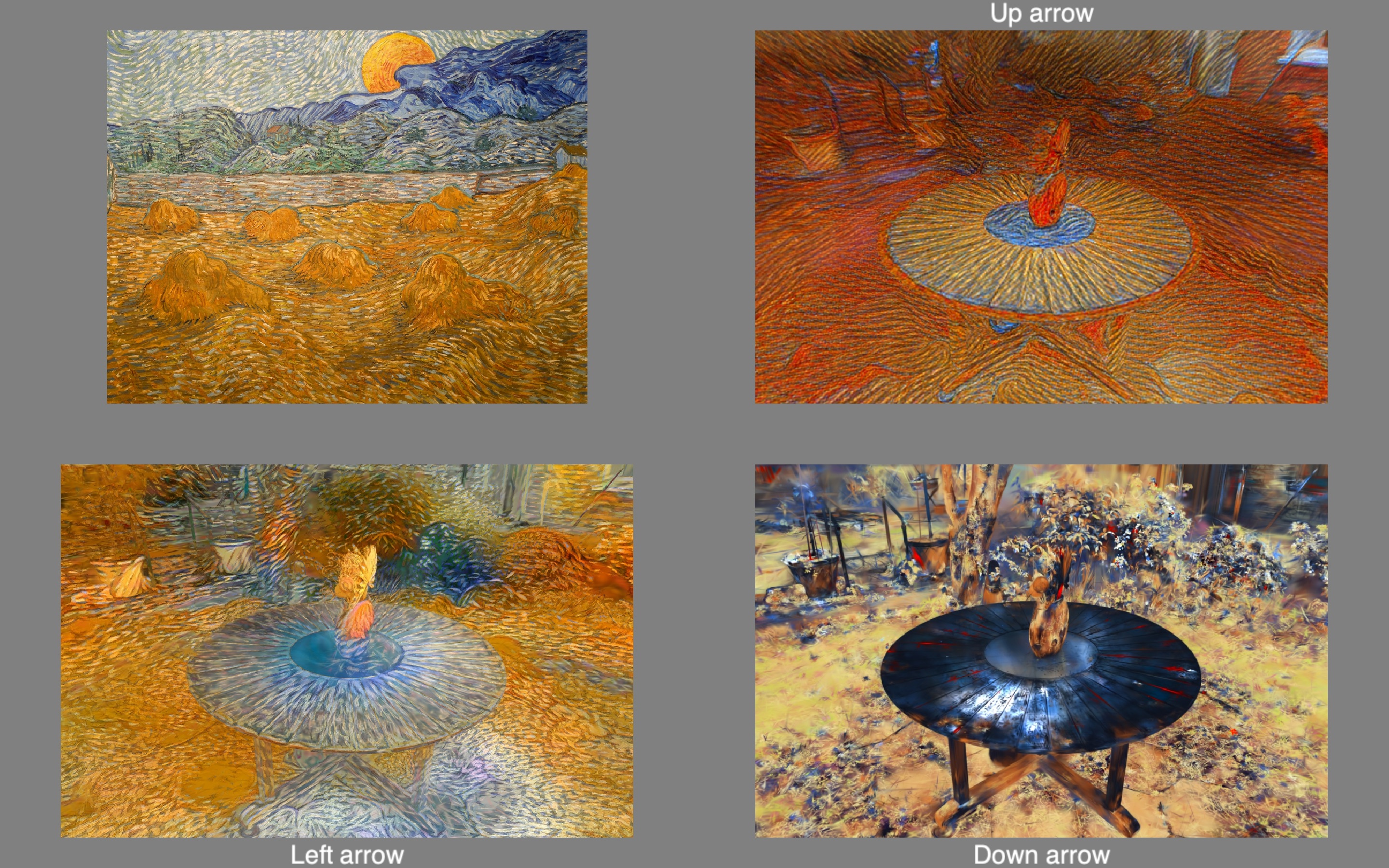}
\caption{Perceptual study. 
The style input image is presented on the top left and the results of each stylization algorithm (SGSST, ARF and StyleGaussian) are presented in a random order. To select the best result, the participant has to press the key indicated next to it.}
\label{fig:perceptual_study_screenshot}
\end{figure*}

\end{document}